\documentclass{article}

\usepackage{PRIMEarxiv}

\usepackage[utf8]{inputenc} 
\usepackage[T1]{fontenc}    
\usepackage[colorlinks = True, linkcolor = violet, citecolor = blue]{hyperref}       
\usepackage{url}            
\usepackage{booktabs}       
\usepackage{amsfonts}       
\usepackage{nicefrac}       
\usepackage{microtype}      
\usepackage{lipsum}
\usepackage{mathtools}
\usepackage{makecell}
\usepackage{stmaryrd}
\usepackage{xcolor}
\usepackage{float}
\usepackage{fancyhdr}       
\usepackage{graphicx}       
\usepackage{wrapfig}

\usepackage[american]{babel}

\usepackage{natbib} 
\usepackage{mathtools} 
\usepackage{tikz} 

\usepackage{amsmath}
\usepackage{amssymb}
\usepackage{mathtools}
\usepackage{makecell}
\usepackage{algorithm}
\usepackage{algorithmic}
\usepackage{enumitem}
\newcommand{\papertitle}{Time-Aware Latent Space Bayesian Optimization}

\newcommand{\x}{\boldsymbol{x}}
\newcommand{\z}{\boldsymbol{z}}

\title{Time-Aware Latent Space Bayesian Optimization}

\author{}

\author{
  Tuan A. Vu\\
  Aalto University \\
  \texttt{tuan.t.vu@aalto.fi} \\
   \And
  Julien Martinelli \\
  Aalto University \\
   \And
  Harri
  Lähdesmäki \\
  Aalto University\\
}

\begin{document}
\maketitle

\begin{abstract}
Latent-space Bayesian optimization (LSBO) extends Bayesian optimization to structured domains, such as molecular design, by searching in the continuous latent space of a generative model.
However, most LSBO methods assume a fixed objective, whereas real design campaigns often face \emph{temporal drift} (e.g., evolving preferences or shifting targets).
Bringing time-varying BO into LSBO is non-trivial: drift can affect not only the surrogate, but also the latent search space geometry induced by the representation.
We propose \emph{Time-Aware Latent-space Bayesian Optimization} (\textsc{TALBO}), which incorporates time in both the surrogate and the learned generative representation via a GP-prior variational autoencoder, yielding a latent space aligned as objectives evolve.
To evaluate time-varying LSBO systematically, we adapt widely used molecular design tasks to drifting multi-property objectives and introduce metrics tailored to changing targets.
Across these benchmarks, \textsc{TALBO} consistently outperforms strong LSBO baselines and remains robust across drift speeds and design choices, while remaining competitive under actually time-invariant objectives.
\end{abstract}

\section{Introduction}\label{sec:intro}
\begin{figure*}
    \centering
    \includegraphics[width=1\linewidth]{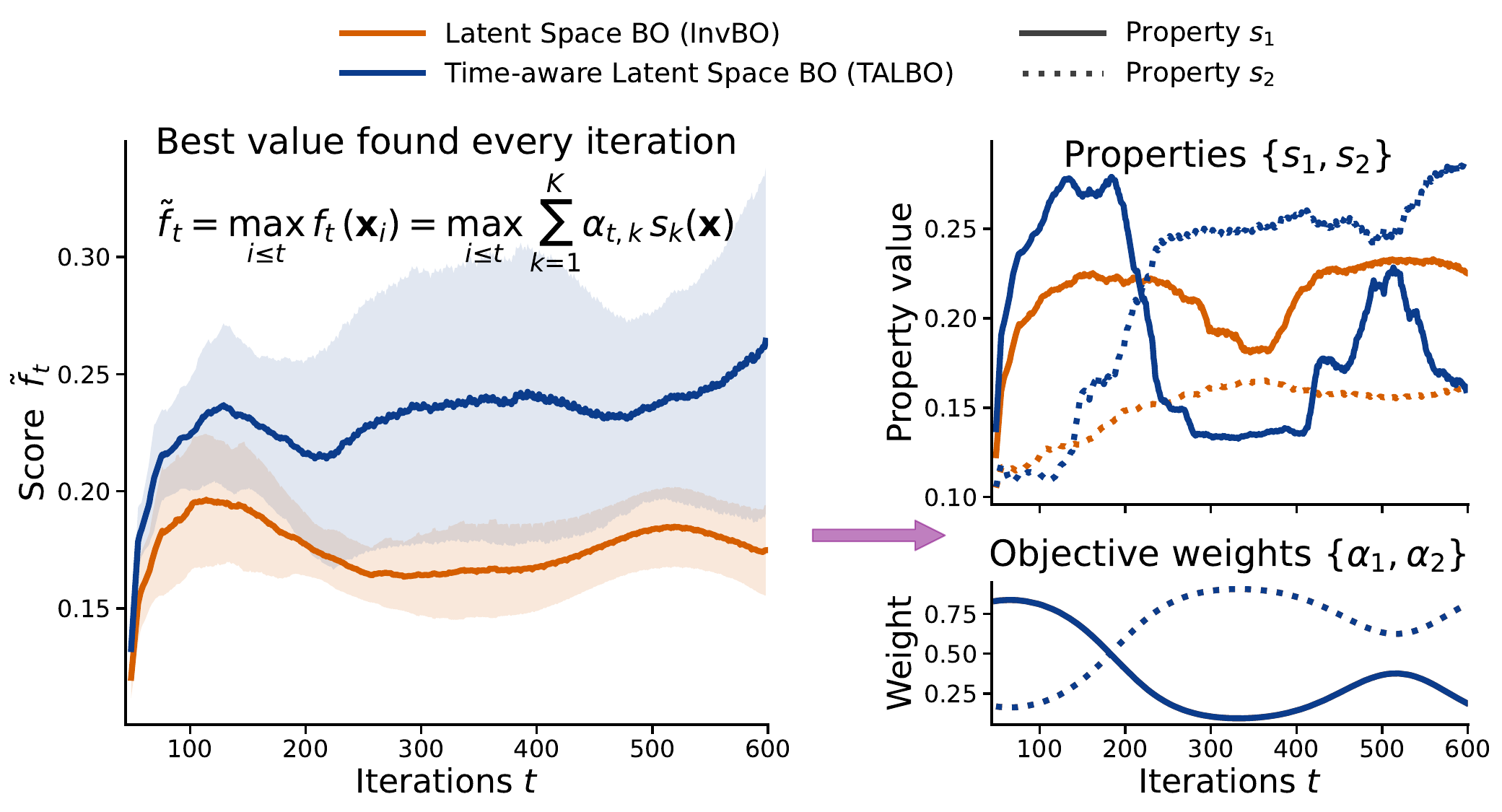}
\caption{\textbf{Overview of our setting.}
At each iteration $t$, a black-box objective $f_t$ is defined as a weighted combination of two molecular properties, with a predefined time-varying weight schedule (details in Section~\ref{sec:mpo}).
Left: The best score under the current objective $f_t$ among all molecules evaluated up to iteration $t$, (mean $\pm$ std over 10 seeds with different initial datasets).
Top-right: mean property components.
Bottom-right: objective weights (identical across seeds).
\textbf{\textsc{TALBO}, our time-aware method adapts more rapidly to changes in objective importance, resulting in consistently higher attained values $\tilde f_t$.}
}

    \label{fig:gist}
\end{figure*}

Many modern design problems reduce to optimizing a black-box function over large, structured spaces. Canonical examples include \emph{de novo} molecular design~\citep{gomez2018automatic}, protein and antibody engineering~\citep{torres2024generative,amin2024bayesian}, and crystal design~\citep{crystalbo}.
In these domains, candidate designs are discrete, high-dimensional objects evaluated through expensive simulations, laboratory assays, or human expertise, making the objective function accessible only via such queries.

To tackle this class of problems, the go-to framework is Bayesian optimization (BO), a sample-efficient approach that maintains a probabilistic surrogate model, typically a Gaussian process (GP,~\citealt{Rasmussen2006}), whose posterior mean and uncertainty are combined through an acquisition function to guide sequential evaluations~\citep{garnett2023bayesian}.
However, vanilla BO methods are not easily amenable to discrete and highly structured domains such as molecules or biological sequences. Latent space BO (LSBO) overcomes this limitation by coupling BO with deep generative models that embed structured objects into a continuous latent space, where optimization is performed before decoding candidates back to the original domain~\citep{gomez2018automatic,tripp2020sample,maus2022local,lee2025latent}.
This formulation 
has become a standard paradigm for \emph{de novo} molecular and protein design.

A key assumption underlying LSBO methods is that the objective function remains fixed throughout optimization.
In many practical design settings, however, objectives evolve over time due to changes in the underlying system or in the optimization criteria itself. 
Examples include antibody discovery against rapidly mutating pathogens, where emerging variants alter antibody binding landscape and designs must be updated or robustified \citep{frei2025deep}; human-in-the-loop optimization \citep{sundin2022human, xu2024principled}; and multi-objective design settings where preferences (scalarization weights) evolve over time \citep{vallerio2015interactive, racz2025changing}.
Failing to account for such temporal variation can result in stale latent representations and surrogate models that no longer reflect the current objective, leading to degraded optimization performance.
While time-varying BO has been studied in continuous domains~\citep{bogunovic2016time,bardou2024too}, a careful treatment of temporal variation in LSBO remains largely unexplored, as the search is mediated by a learned generative representation.

\textbf{Contributions.}
We propose \emph{Time-Aware Latent-space Bayesian Optimization} (\textsc{TALBO}), a framework for time-varying latent space optimization that models temporal drift \emph{both} in the surrogate and in the learned representation.
Figure~\ref{fig:gist} provides a schematic overview of the setting and highlights why time-awareness is essential for tracking evolving objectives in structured design.
Our contributions are:
\begin{itemize}[leftmargin=9mm]
  \item \textbf{Conceptual.} We formalize \emph{time-varying LSBO}: an optimization setting where the objective changes across rounds,  while search is performed in a learned latent space.
  This formulation exposes a key challenge absent from standard LSBO---\emph{time can enter both the objective model and the latent embedding}---and motivates principled mechanisms to handle temporal drift.

  \item \textbf{Methodological.} We introduce \textsc{TALBO}, which couples a spatio-temporal surrogate with a conditional generative representation guided by time and possible other covariates, aligning the latent search space with the current objective function.

 \item \textbf{Empirical.} We adapt widely used molecular design benchmarks to a controlled time-varying regime and perform extensive experiments under fixed budgets.
  \textsc{TALBO} consistently outperforms strong LSBO baselines, and a broad suite of controlled studies  shows that these gains are robust across design choices and experimental settings.
\end{itemize}

\section{Preliminaries}

Let $\mathcal{X}$ be an input space and $\mathcal{Y}\subset\mathbb{R}$ an output space. 
We study \emph{time-varying} black-box optimization, where at each iteration $t\in\{1,\dots,T\}$
the learner observes noisy evaluations of an objective $f_t:\mathcal{X}\to\mathcal{Y}$,
\begin{equation}
y_{tb} = f_t(\x_{tb}) + \varepsilon_t,\qquad \varepsilon_t\sim\mathcal{N}(0,\sigma^2),
\end{equation}
for a batch $\{\x_{t1},\ldots,\x_{tB}\}  \in  \mathcal{X}^B$, and aims to select a query  $\x_{t+1}$ (or batch) that performs well under the evolving objective. For methodological descriptions, we may assume $B=1$.
Since $\arg\max_{\x\in\mathcal{X}} f_t(\x)$ may change with $t$, the goal is not to recover a single static maximizer, but to \emph{adaptively track} high-value inputs as $f_t$ drifts.

$\mathcal{X}$ may be discrete, high-dimensional, and structured (e.g., molecules), making optimization difficult.
This poses two challenges: handling time-varying objectives and navigating complex search spaces. Next, we introduce the components used to address them: time-varying Bayesian optimization and learned representations for structured designs.

\subsection{Time-Varying Bayesian Optimization}

BO provides a sequential strategy for black-box optimization and can be extended to time-varying objectives.

\textbf{Bayesian Optimization.}
Given $\mathcal{D}_t=\{(\x_i,t_i,y_i)\}_{i=1}^t$, BO maintains a \emph{probabilistic surrogate} over the objective, denoted $p(f \mid \mathcal{D}_t)$, which induces a predictive distribution for $f(\x)$. 
The next query is selected by maximizing an \emph{acquisition function} that depends on this surrogate,
\begin{equation}
\x_{t+1}\in\arg\max_{\x\in\mathcal{X}}
\alpha\bigl(\x;\, p(f \mid \mathcal{D}_t)\bigr),
\end{equation}
where $\alpha$ trades off predicted value (exploitation) and uncertainty (exploration) under the surrogate.

\textbf{Gaussian Processes.}
A common choice for the probabilistic surrogate is a Gaussian process (GP),
\begin{equation}
f \sim \mathcal{GP}\!\bigl(\mu(\cdot), k(\cdot,\cdot)\bigr),
\end{equation}
where $\mu$ is typically set to zero and $k$ is a positive semidefinite kernel encoding prior smoothness.
Assuming a Gaussian likelihood and fixed hyperparameters,
conditioning on $\mathcal{D}_t$ yields a Gaussian predictive distribution
for $f(\x)$ with closed-form mean and variance.
However, as exact inference scales cubically in $|\mathcal{D}_t|$,
sparse or variational approximations are often used in practice~\cite{titsias2009variational}.

\textbf{Time-Varying BO.}
Here, the objective changes across rounds, $f_t(\x)$, and the surrogate must therefore model temporal evolution and modulate the relevance of past observations. 
A common approach is to place a GP prior over a spatio-temporal function $f_t(\x) := f(\x,t)$, using a kernel defined on $(\x,t)$, such as
$
k\big((\x,t),(\x',t')\big):=k_x(\x,\x')\,k_t(t,t').
$
The acquisition function at round $t$ is computed from the posterior predictive distribution of the current round $f(\cdot,t)$~\citep{bogunovic2016time}.

\textbf{Metrics.}
As the objective changes over time, standard best-so-far metrics are insufficient.
We therefore consider two evaluation criteria. (i) \emph{Current-objective best-so-far value},
\begin{equation}
    \tilde{f_t} = \max_{i \le t} f_t(\x_i)
\label{eq:bpt}
\end{equation}
which evaluates all previously queried designs under the
current objective $f_t$ and reports the best achievable value at time $t$.
(ii) The \emph{cumulative regret},
\begin{equation}
R_T := \sum_{t=1}^T \bigl(f_t^\star - \tilde f_t \bigr),
\label{eq:cr}
\end{equation}
where $f_t^\star := \max_{\x \in \mathcal{X}} f_t(\x)$
denotes the optimal value of the objective at time $t$.
It is worth noticing that $\tilde f_t$ is generally \emph{not} monotone and can decrease when the objective shifts and previously good designs become less aligned with the new trade-off, as displayed in Figure~\ref{fig:gist}.
\subsection{Latent Space Bayesian Optimization}

When $\mathcal{X}$ is discrete, structured, or high-dimensional,
direct optimization is challenging.
LSBO addresses this by learning a continuous representation,
for example via a Variational Autoencoder (VAE)~\citep{kingma2013auto}.

A VAE defines a probabilistic encoder $q_\phi(\z \mid \x)$
and decoder $p_\theta(\x \mid \z)$.
In practice, LSBO operates on a deterministic embedding,
e.g., the posterior mean $\mathbb{E}_{q_\phi(\z \mid \x)}[\z]$,
together with a decoding map
\begin{equation}
\boldsymbol{\Gamma}(\z)
:= \arg\max_{\x} p_\theta(\x \mid \z),
\end{equation}
which maps latent points back to the design space.
This induces a continuous latent space
$\mathcal{Z} \subset \mathbb{R}^d$
in which Bayesian optimization can be performed:
\begin{equation}
\z_{t+1}
\in
\arg\max_{\z \in \mathcal{Z}}
\alpha\bigl(\z; p(g \mid \mathcal{D}^{z}_t)\bigr),
\end{equation}
where
\begin{equation}
g(\z) := f(\boldsymbol{\Gamma}(\z)),
\qquad
\mathcal{D}^{z}_t = \{(\z_i, t_i, y_i)\}_{i=1}^t.
\end{equation}
The selected latent point is decoded as
\begin{equation}
\x_{t+1} = \boldsymbol{\Gamma}(\z_{t+1}).
\end{equation}
\textbf{Representation updates.}
Vanilla LSBO keeps the generative model fixed during optimization.
Some variants periodically retrain it, often jointly with the surrogate~\citep{chu2024inversion, maus2022local},
to better align latent geometry with observed values.
These approaches, however, assume a static objective
and do not address temporal drift.

\section{Problem Statement}

Extending LSBO to time-varying objectives introduces a structural challenge.
In LSBO, optimization occurs in a learned latent space $\mathcal{Z}$
whose geometry is induced by a generative model.
Under temporal drift, a static latent representation
may become misaligned with the evolving objective.

Classical time-varying BO addresses drift at the surrogate level
by modeling $f(\x,t)$ with a spatio-temporal kernel.
In LSBO, however, the surrogate operates on latent codes via
\begin{equation}
g(\z,t) := f_t(\boldsymbol{\Gamma}(\z)),
\end{equation}
where $\boldsymbol{\Gamma}$ denotes the decoder.
Modeling time only in the surrogate can therefore be insufficient
if the latent geometry itself remains static.

\textbf{Dual Temporal Modeling.}
We incorporate time at two levels:
a spatio-temporal GP models $g(\z,t)$ in latent space,
while the mapping $\x \mapsto \z$
is endowed with a GP prior over latent functions.
This joint treatment allows both uncertainty and latent geometry
to adapt under temporal drift.

\section{Time-Aware LSBO with a GP-Prior Generative Model}

We propose \emph{Time-Aware Latent Space Bayesian Optimization} (\textsc{TALBO}), which instantiates dual temporal modeling by leveraging the GP-prior VAE with scalable basis-function approximation and global variational parameters (DGBFGP)~\citep{balk2025bayesian}. 
To our knowledge, this is the first use of temporal deep latent variable models within BO.
Section~\ref{sec:dgbfgp} presents the latent representation,
Section~\ref{sec:dgbfgp_vi} the objective,
and Section~\ref{sec:dgbfgp_time} the treatment of time.
Algorithm~\ref{alg:tvlsbo_dgbfgp} summarizes the full \textsc{TALBO} loop.

\subsection{Structured latent representation}
\label{sec:dgbfgp}

Each design $\x \in \mathcal{X}$ is associated with $R$ auxiliary covariates
$\mathbf{c}(\x,t)\in\mathbb{R}^{R}$, which may include continuous descriptors and time.
Modeling these covariates explicitly enables the latent geometry
to capture structured and temporally coherent variation,
which is essential under objective drift.
Given covariates $\mathbf{c} = (c^{(1)},\ldots,c^{(R)}) := \mathbf{c}(\x,t)$, DGBFGP defines a latent code $\z(\mathbf{c})\in\mathbb{R}^{d}$
and a decoder $p_\theta(\x \mid \z)$.
The latent geometry is induced by an additive Gaussian process prior
over covariate-dependent latent functions.

\textbf{Basis-function GP prior.}
For each covariate component $c^{(r)}$,
we place a GP prior on a univariate latent function
$h^{(r)}$,
\begin{equation}
h^{(r)}(c^{(r)}) \sim \mathcal{GP}\!\bigl(0, k^{(r)}(\cdot,\cdot)\bigr),
\end{equation}
where $k^{(r)}$ is a covariance kernel with hyperparameters
$\sigma_r,\ell_r$.
To obtain a scalable parameterization,
DGBFGP employs a Hilbert-space basis-function approximation~\citep{solin2020hilbert},
yielding the linear form
\begin{align}
h^{(r)}(c^{(r)}) &\approx \boldsymbol{a}^{(r)\top}\boldsymbol{\phi}^{(r)}(c^{(r)}), \\
\boldsymbol{a}^{(r)} &\sim \mathcal{N}\!\bigl(\mathbf{0}, \boldsymbol{S}^{(r)}(\sigma_r,\ell_r)\bigr),
\end{align}
where $\boldsymbol{\phi}^{(r)}(\cdot) \in \mathbb{R}^{M}$ denotes Hilbert features
and $\boldsymbol{S}^{(r)}(\sigma_r,\ell_r)$ is diagonal.
We use squared-exponential kernels for continuous covariates.
Detailed Hilbert-space feature constructions are provided in Appendix~\ref{sec:dgbfgp_full}.

GP prior VAE models assume independent GP priors for each of the latent dimensions~\citep{casale2018gaussian}. Defining $\boldsymbol{A}^{(r)} = [\boldsymbol{a}^{(r,1)}{}^\top,\ldots,\boldsymbol{a}^{(r,d)}{}^\top]^\top \in \mathbb{R}^{d \times M}$, the basis function approximation for the GP prior across all latent dimensions is
\begin{align}
    \z(\mathbf{c}) &= \boldsymbol{A}^{(r)} \boldsymbol{\phi}^{(r)}(c^{(r)}) \\
    \boldsymbol{a}^{(r,l)} &\sim \mathcal{N}\!\bigl(\mathbf{0}, \boldsymbol{S}^{(r)}(\sigma_r,\ell_r)\bigr),
\end{align}
where we have assumed shared kernel hyperparameters across latent dimensions.

\textbf{Additive latent code.}
DGBFGP represents the latent code as a sum of covariate effects:
\begin{equation}
\z(\mathbf{c};\boldsymbol{A})
=
\sum_{r=1}^{R} \boldsymbol{A}^{(r)} \boldsymbol{\phi}^{(r)}(c^{(r)}),
\qquad
\boldsymbol{A}^{(r)} \in \mathbb{R}^{d \times M},
\end{equation}
where $\boldsymbol{A} = [\boldsymbol{A}^{(1)},\ldots,\boldsymbol{A}^{(R)}]$. DGBFGP enforces smoothness with appropriate priors for the kernel hyperparameters:
\begin{align}
\sigma_r,\ell_r &\sim \mathrm{Lognormal}(0,1).
\end{align}
Note that time is included as one of the covariates,
making the latent geometry explicitly time-dependent.

\subsection{Variational learning}
\label{sec:dgbfgp_vi}

Given a training set $\{(\x_n,\mathbf{c}_n)\}_{n=1}^{N}$,
we model designs through the decoder likelihood
\begin{equation}
p_\theta(\x_n \mid \boldsymbol{A},\mathbf{c}_n) := p_\theta\!\bigl(\x_n \mid \z(\mathbf{c}_n;\boldsymbol{A})\bigr).
\end{equation}
We learn global parameters with a factorized variational posterior
\begin{equation}
q(\boldsymbol{A},\sigma,\ell) = q(\boldsymbol{A})\,q(\sigma)\,q(\ell),
\end{equation}
where $q(\boldsymbol{A})$ is Gaussian and $q(\sigma)$ and $q(\ell)$ are log-normal.
The ELBO takes the form
\begin{equation}
\begin{aligned}
\hspace{-3pt}\mathcal{L}_{\mathrm{DGBFGP}}
&=
\mathbb{E}_{q(\boldsymbol{A})}\!\Big[\sum_{n=1}^{N}\log p_\theta\!\bigl(\x_n \mid \z(\mathbf{c}_n;\boldsymbol{A})\bigr)\Big]
\\& \quad -
\mathbb{E}_{q(\sigma)q(\ell)}\!\Big[\mathrm{KL}\!\big(q(\boldsymbol{A})\,\|\,p(\boldsymbol{A}\mid\sigma,\ell)\big)\Big] \\
&
\quad -\mathrm{KL}\!\big(q(\sigma)\,\|\,p(\sigma)\big)
-\mathrm{KL}\!\big(q(\ell)\,\|\,p(\ell)\big),
\end{aligned}
\label{eq:finalloss}
\end{equation}
and is optimized by stochastic gradient methods with mini-batching~\citep{balk2025bayesian}.

In addition to the variational objective in Equation~\eqref{eq:finalloss}, 
our time-aware representation updates incorporate two complementary mechanisms. 
First, we add alignment regularizers that encourage the latent geometry to remain smooth 
and well-scaled with respect to observed objective values~\citep{lee2023advancing}. 
Second, we apply latent inversion to mitigate misalignment between evaluated designs 
and their latent representations~\citep{chu2024inversion}.
Finally, we train a variational GP surrogate on latent observations, using an SVGP objective~\citep{hensman2015scalable}.
These components are detailed in Appendix~\ref{app:additional-loss-terms},~\ref{app:invbo}, and~\ref{app:svgp_loss}, respectively.

\subsection{Time-Conditioned Latent Representation}
\label{sec:dgbfgp_time}

Since time $t$ is included as an additional continuous covariate in $\mathbf{c}(\x,t)$,
the latent code $\z(\mathbf{c}(\x,t);\boldsymbol{A})$ becomes time-dependent,
inducing a temporally coherent latent geometry.
This allows the representation to evolve as objectives drift.

Algorithm~\ref{alg:tvlsbo_dgbfgp} summarizes the resulting BO loop.
At each iteration, we embed evaluated design with DGBFGP covariates, update a spatio-temporal surrogate over $(\z,t)$,
and select a latent query to decode and evaluate.
The generative model may be updated during optimization; when it is, latent embeddings are recomputed to maintain alignment. Of note, DGBFGP is decoder-only, so the posterior mean embedding of design $\x_i$ is computed as $\z_i = \mathbb{E}_{q_t(\boldsymbol{A})}\!\big[ \z(\mathbf{c}(\x_i,t_i);\boldsymbol{A}) \big]$.

Most LSBO papers describe only the high-level loop and omit many practical details; here we make a best effort to explicitly enumerate the modules and steps needed for faithful reproduction.
A detailed procedure, including the update trigger, alignment regularizers~\citep{lee2023advancing}, and latent inversion~\citep{chu2024inversion}, is provided in Algorithm~\ref{alg:talbolong}.

\begin{algorithm}[t]
\caption{\textsc{TALBO} (simplified version)}
\label{alg:tvlsbo_dgbfgp}
\begin{algorithmic}[1]
\REQUIRE Initial evaluated set $\{(\x_i,t_i,y_i)\}_{i=1}^{n_0}$.
\STATE Train DGBFGP on initial corpus to obtain $q_0(\boldsymbol{A},\sigma,\ell)$.
\STATE Construct initial BO dataset\\
$\mathcal{D}_0  = \big\{ \big( \z_i, t_i, y_i \big) \big\}_{i=1}^{n_0}, 
\z_i = \mathbb{E}_{q_0}\!\big[ \z(\mathbf{c}(\x_i,t_i);\boldsymbol{A}) \big]$.
\FOR{$t = 1$ \TO $T$ or until criterion satisfied}
    \STATE Fit spatio-temporal surrogate on $\mathcal{D}_{t-1}$ (Eq.~\ref{eq:svgp_elbo}).
    \STATE Select $\hat{\z} \in \arg\max_{\z \in \mathcal{Z}} \alpha\!\big(\z; p(g(\cdot,t)\mid\mathcal{D}_{t-1})\big)$.
    \STATE Decode design $\x_t \sim p_{\theta_{t-1}}(\cdot \mid \hat{\z})$.
    \STATE Observe $y_t = f_t(\x_t) + \varepsilon_t$ and  $\mathbf{c}_t = \mathbf{c}(\x_t,t)$.
    \STATE Compute embedding $\z_t \gets \mathbb{E}_{q_{t-1}}[\z(\mathbf{c}_t;\boldsymbol{A})]$.
    \STATE Update BO dataset $\mathcal{D}_t \gets \mathcal{D}_{t-1} \cup \{(\z_t,t,y_t)\}$.
    \IF{representation update triggered (Alg.~\ref{alg:talbolong}-L18)}
        \STATE Update DGBFGP posterior and decoder params.\ $(q_t,\theta_t)$ via warm-start SVI on current data.
        \STATE Recompute $\mathcal{D}_t$ by re-embedding designs via $q_t$.
    \ENDIF
\ENDFOR
\end{algorithmic}
\end{algorithm}

\section{Related work}
\textbf{Latent space Bayesian optimization}~\citep{gomez2018automatic}
extends traditional BO to discrete, structured,
or high-dimensional input spaces by leveraging generative models
such as VAEs~\citep{kingma2013auto}
to embed structured designs into a continuous latent space
where standard BO can be applied.
However, performing BO over a fixed VAE representation can suffer from
poor sample efficiency, as the latent geometry may be misaligned
with the optimization objective.
To mitigate this, recent work has explored adapting the latent space
during optimization, including weighted retraining schemes~\citep{tripp2020sample},
metric learning objectives~\citep{grosnit2021high},
and joint inference over the surrogate and generative mapping, as in LOL-BO~\citep{maus2022local}.
CoBO~\citep{lee2023advancing} proposes a Lipschitz-constrained fine-tuning objective
to better align the latent representation with the target function,
and InvBO~\citep{chu2024inversion} extends this line of work to mitigate decoder-induced
misalignment via latent inversion.
Other approaches decouple the generative and surrogate components~\citep{cowboys25}
and incorporate domain-specific kernels such as the Tanimoto kernel~\citep{tripp2023tanimoto}.
See~\cite{gonzalez-duque2024a} for a comprehensive survey of LSBO methods.
Despite this extensive literature,
the extension to time-varying objectives remains underexplored.
In such settings, the latent geometry may evolve not only through representation updates,
but also due to temporal drift in the objective itself.

\textbf{Time-varying Bayesian optimization} tackles settings where the objective evolves over time. A common approach incorporates time as an additional input to a spatio-temporal surrogate kernel~\citep{marchant2012bayesian, marchant2014sequential}.
Subsequent work has developed methods that manage outdated observations either by periodically resetting the dataset or by gradually discounting past data~\citep{bogunovic2016time, brunzema2022controller}.
More recently, adaptive relevance measures based on distributional distances have been proposed to assess the influence of past observations on future predictions~\citep{bardou2024too}.
Despite these advances, time-varying BO has been studied almost exclusively in the original input space. Integrating it with latent-space optimization introduces additional challenges:
in adaptive LSBO, the latent geometry evolves through representation updates,
and in time-varying settings it may also shift due to temporal drift in the objective.
Our method accounts for both sources of evolution.

\textbf{Generative backbones for latent representations} started with VAEs in LSBO~\citep{kingma2013auto, gomez2018automatic}.
Extensions replace the standard Gaussian prior with Gaussian process priors
to encode temporal or covariate-dependent correlations in latent space,
as in GPPVAE~\citep{casale2018gaussian}, GP-VAE~\citep{fortuin2020gp} and additive GP-prior models~\citep{ramchandran2021longitudinal}, with BO applications~\citep{ramchandran2025highdimensional}. See Appendix~\ref{app:vae_background} for more background.
Recent work, like DGBFGP~\citep{balk2025bayesian}, improves scalability, with basis-function approximations and global parameterizations.
Alternative backbones have also been explored,
like normalizing flows to obtain invertible latent mappings and
mitigate reconstruction-induced discrepancies~\citep{lee2025latent}.

\section{Experiments}

\textbf{Baselines.}
We compare \textsc{TALBO} against representative LSBO methods: InvBO~\citep{chu2024inversion}, CoBO~\citep{lee2023advancing}, and LOL-BO~\citep{maus2022local}. 
\textsc{TALBO} and these baselines are equipped with Trust Region Bayesian Optimization (TuRBO;~\citealt{eriksson2019scalable}), which improves high-dimensional BO by restricting search to adaptive local trust regions centered at promising incumbents.
In addition, we report three baselines: standard LSBO without trust-region constraints, and LSBO combined with TuRBO, and a random search that samples latent codes uniformly and decodes them to candidate molecules.
Our method is implemented with both the SELFIES VAE of~\cite{maus2022local} and our proposed SELFIES DGBFGP model.
Further implementation details are provided in Appendix~\ref{app:expdetails}.

\textbf{Implementation details.}
We follow~\cite{chu2024inversion} and employ Thompson sampling as the acquisition function, using a variational GP surrogate~\citep{hensman2015scalable} with a deep kernel~\citep{wilson2016deep} and an RBF covariance function (Appendix~\ref{app:svgp_loss}).
The surrogate and acquisition components are implemented in \texttt{BoTorch}~\citep{balandat2020botorch}. 
For molecular discovery, we consider both the SELFIES VAE of~\cite{maus2022local} and our proposed SELFIES DGBFGP model, each pretrained on the 1.27M unlabeled molecules from the GuacaMol dataset~\citep{brown2019guacamol}. Code will be available upon acceptance.

\subsection{Multi-property optimization with drifting priorities}\label{sec:mpo}

We consider a controlled scenario in which the relative importance
of $K$ property scores changes over time.
This setting mimics situations where preferences shift,
leading to a time-varying objective.

Let $s_k(\mathbf{x}) \in [0,1]$ denote $K$ base property scores.
At time $t$, the objective is defined as a convex combination
\begin{equation}
    f_t(\mathbf{x})
    =
    \sum_{k=1}^K \alpha_{t,k} \, s_k(\mathbf{x}),
    \label{eq:mpo}
\end{equation}
where
$\boldsymbol{\alpha}_t \in \mathbb{R}_+^K$
lies on the simplex,
$\sum_{k=1}^K \alpha_{t,k} = 1$.

\textbf{Synthetic weight generation $\boldsymbol{\alpha}$.}
To induce smoothly drifting priorities,
we sample $K$ independent latent functions
\begin{equation}
    w_k(t) \sim \mathcal{GP}\!\bigl(0, k_{w}(t,t')\bigr),
    \qquad k=1,\dots,K,
\label{eq:weight}
\end{equation}
where $k_{w}$ is a temporal covariance kernel.
We then define
\begin{equation}
    \alpha_{t,k}
    =
    \frac{\exp\big(\beta w_k(t)\big)}
    {\sum_{j=1}^K \exp\big(\beta w_j(t)\big)},
    \quad k=1,\dots,K,
\label{eq:softmax}
\end{equation}
which maps the latent GP draws to the probability simplex.
The temperature parameter $\beta>0$ controls
how sharply priorities concentrate over time (Appendix~\ref{app:hyperparameters}).

\textbf{Property scores $s$.}
To create tasks that require explicit trade-offs between components
as their relative importance shifts, we construct objectives
based on molecular similarity scores.
Following the median molecule tasks introduced by~\cite{brown2019guacamol},
we select four scoring functions derived from the Guacamol benchmark suite.
Specifically, two tasks correspond to components of the
\emph{Median Molecules 1} and \emph{Median Molecules 2} benchmarks,
while the remaining two are adapted from the similarity
and rediscovery tasks.
Each base score $s_k(\x)\in[0,1]$ and $f_t(\x)=\sum_{k=1}^K \alpha_{t,k}s_k(\x)$, hence $f_t(\x)\in[0,1]$ and $f_t^\star=1$ for all $t$.
The full set of scoring functions is detailed in Table~\ref{tab:tasks_scoring_functions}.

\textbf{Experimental protocol and budget.}
Each run consists of $T=600$ BO rounds.
Before the BO loop begins, we construct an initial dataset of $N_{\mathrm{init}}=100$ evaluations by sampling 100 molecules uniformly from the GuacaMol pre-training corpus and evaluating each of them under the first 50 iterations. 
These evaluations are shared across methods for each seed, but time-agnostic baselines are only provided with the same 100 molecules evaluated solely at the most recent time step.
Then, at round $t$, a batch of $B=10$ latent points is selected (via $q$-batch Thompson sampling), decoded to molecules, and evaluated under $f_t$.
\textsc{TALBO} uses covariates described in Appendix~\ref{sec:dgbfgp_full}.

\begin{table}[t]
\centering
\small
\begin{tabular}{ll}
\toprule
\textbf{Task} & \textbf{Scoring functions} \\
\midrule
Median molecules 1 &
\makecell[l]{
$\mathrm{sim}(\text{camphor}, \mathrm{ECFC4})$ \\
$\mathrm{sim}(\text{menthol}, \mathrm{ECFC4})$
} \\
\midrule
Median molecules 2 &
\makecell[l]{
$\mathrm{sim}(\text{tadalafil}, \mathrm{ECFC6})$ \\
$\mathrm{sim}(\text{sildenafil}, \mathrm{ECFC6})$
} \\
\midrule
Composite Similarity 1 &
\makecell[l]{
$\mathrm{sim}(\text{Sitagliptin}, \mathrm{ECFC4})$ \\
$\mathrm{sim}(\text{Ranolazine}, \mathrm{AP})$ \\
$\mathrm{sim}(\text{Osimertinib}, \mathrm{FCFC4})$ 
} \\
\midrule
Composite Similarity 2 &
\makecell[l]{
$\mathrm{sim}(\text{Celecoxib}, \mathrm{ECFC4})$ \\
$\mathrm{sim}(\text{Troglitazone}, \mathrm{ECFC4})$ \\
$\mathrm{sim}(\text{Thiothixene}, \mathrm{ECFC4})$
} \\
\bottomrule
\end{tabular}
\caption{Tasks and their associated scoring functions.}
\label{tab:tasks_scoring_functions}
\end{table}

\begin{table}
\small
\centering
\begin{tabular}{lccc}
\toprule
Baseline  & $\ell_w=0.1$ & $\ell_w=0.2$ & $\ell_w=0.3$ \\
\midrule
\textsc{TALBO} & \textbf{1.98 $\pm$ 1.23} & \textbf{1.75 $\pm$ 1.08} & \textbf{1.90 $\pm$ 1.15} \\
InvBO & 3.77 $\pm$ 1.39 & 3.90 $\pm$ 1.24 & 3.52 $\pm$ 1.38 \\
CoBO & 2.48 $\pm$ 1.04 & \underline{2.45 $\pm$ 1.30} & \underline{2.30 $\pm$ 1.16} \\
LOL-BO & \underline{2.42 $\pm$ 1.01} & 2.73 $\pm$ 1.11 & 2.92 $\pm$ 1.10 \\
LS-BO & 6.41 $\pm$ 0.65 & 6.62 $\pm$ 0.60 & 6.59 $\pm$ 0.65 \\
TurBO & 4.76 $\pm$ 1.06 & 4.76 $\pm$ 1.19 & 4.80 $\pm$ 1.07 \\
Random & 5.35 $\pm$ 0.81 & 5.79 $\pm$ 0.97 & 5.62 $\pm$ 0.89 \\
\bottomrule
\end{tabular}
\caption{
\textbf{Ablation study.} Final baseline rank (mean $\pm \tfrac{1}{2}$ std over all tasks and 10 random seeds) for different GP lengthscales $\ell_{w} \in \{0.1, 0.2, 0.3\}$ (Equation~\ref{eq:weight}). 
Ranks are computed per task and seed from cumulative regret (lower is better) and then averaged across tasks and seeds. 
Bold entries denote the best (lowest) average rank for each $\ell_{w}$, and underlined entries the second-best. 
\textbf{\textsc{TALBO} achieves the best average rank across all tested lengthscales, demonstrating robustness to varying rates of objective drift.}
}
\label{tab:lengthscale}
\end{table}

\subsection{Results}

\begin{figure*}[t]
    \centering
    \includegraphics[width=.95\linewidth]{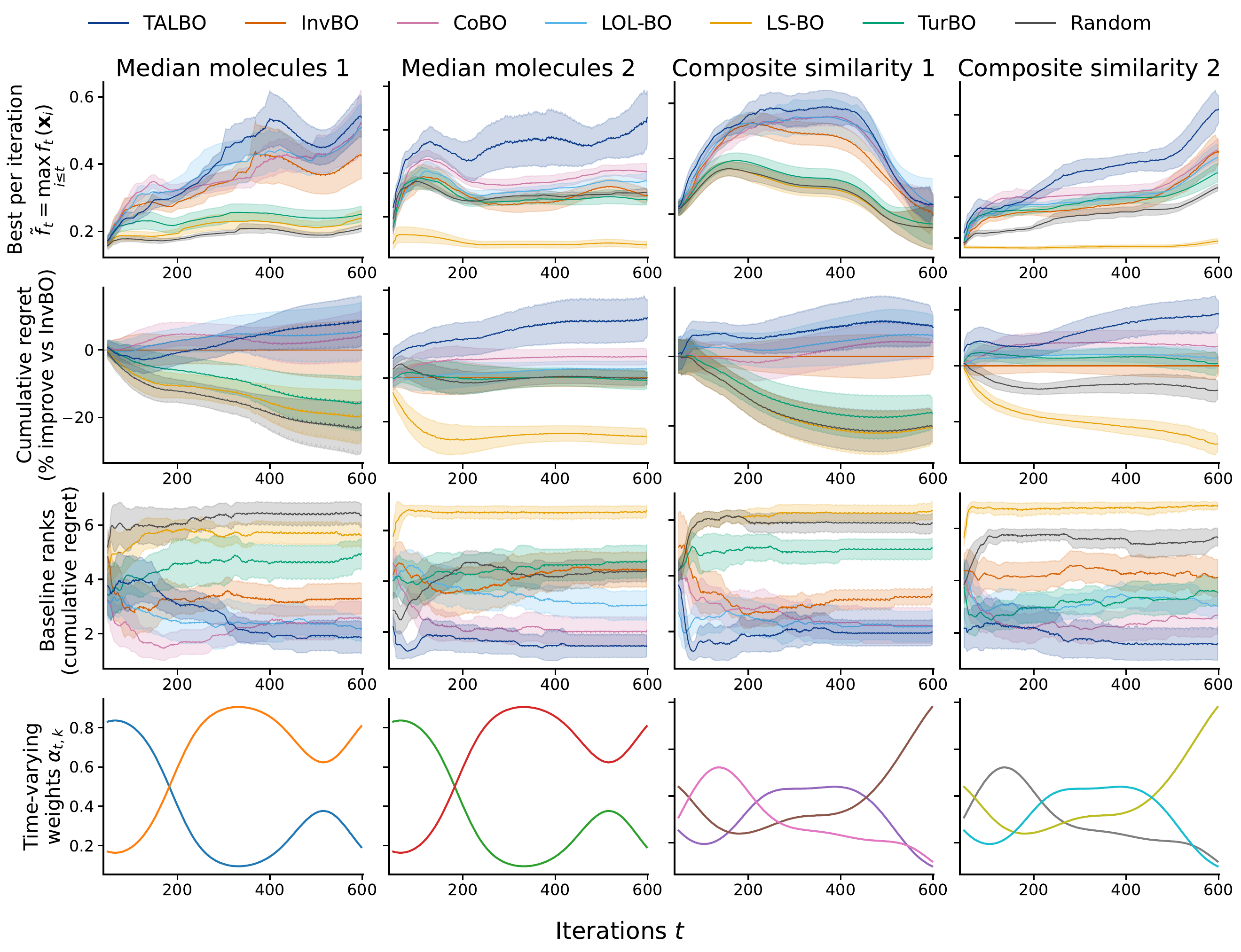}
\caption{
\textbf{Performance comparison across dynamic multi-objective tasks.}
First row: current-objective best-so-far value.
Second row: cumulative regret, expressed as percentage improvement relative to \textsc{InvBO}.
Third row: baseline rank at each iteration based on cumulative regret (lower is better).
Fourth row: time-varying objective weights governing the dynamic multi-objective trade-off, using $\ell_w = 0.2$ for $k_w$ (Equation~\ref{eq:weight}).
Mean $\pm \tfrac{1}{2}$std over 10 random seeds.
All trajectories are smoothed using a moving average for improved readability.
\textbf{Across all tasks, \textsc{TALBO} consistently outperforms other baselines, attaining higher best-found values, lower regret, and better average rank.}
}
    \label{fig:results}
\end{figure*}

\textbf{Overall performance.}
Figure~\ref{fig:results} compares all methods using three complementary views: the current-objective best-so-far value (Equation~\ref{eq:bpt}), its associated cumulative regret (Equation~\ref{eq:cr}), and regret-based ranks.
Across all four tasks (Table~\ref{tab:tasks_scoring_functions}), \textsc{TALBO} achieves the strongest overall performance in terms of best-per-time (first row), attaining the highest values over most of the optimization horizon.
Competing baselines can occasionally match or even exceed \textsc{TALBO} early on, but they typically plateau or adapt more slowly once the objective priorities shift.
The same trend is reflected in cumulative regret (second row), reported as percentage improvement relative to InvBO for readability.
Across most tasks, \textsc{TALBO} attains the most favorable regret profiles, with an advantage that typically grows as optimization proceeds.
An exception is the first task (first column), where the regret curves are closer and competing baselines remain competitive for larger portions of the horizon.
To summarize performance in a scale-free way, the third row reports ranks induced by cumulative regret (lower is better).
This view is helpful when absolute gaps in best-per-time or regret are small and hard to interpret, since it highlights consistent relative performance over time and across tasks.
Overall, all three metrics support the same conclusion: explicit time-awareness improves tracking of drifting objectives over strong LSBO baselines.
For completeness, Figure~\ref{fig:instant} reports the results using the instantaneous objective value $f_t(\mathbf{x}_t)$ at each iteration, leading to consistent qualitative conclusions.

\textbf{Relating performance to the time-varying weights.}
The last row of Figure~\ref{fig:results} shows the time-varying objective weights $\alpha_{t,k}$ that define the scalarization.
These exhibit clear regime changes (e.g., crossings between components and periods where one component dominates), which induce changes in which regions of chemical space are optimal under $f_t$.
We observe that many turning points and transient drops in $\tilde f_t$ coincide with weight transitions: when priorities shift, previously best molecules are re-scored under the new mixture and can temporarily become suboptimal, causing $\tilde f_t$ to decrease.
We hypothesize that \textsc{TALBO}'s time-aware surrogate and generative model help it refocus more quickly after such shifts, sustaining higher best-per-time values.

\textbf{How hard is it to beat the pre-training set?}
To contextualize the difficulty of these time-varying objectives, we compute reference curves directly from the pre-training corpus.
At each time $t$, we evaluate $f_t$ over the full pre-training set (1.27M GuacaMol molecules) and report the dataset maximum (solid black) together with high quantiles (dashed; top 5\%, 1\%, 0.1\%, and 0.01\%, see Figure~\ref{fig:mainwdataset}).
These curves indicate what could be achieved at each iteration by exhaustively searching the pre-training set.
Complementarily, Figure~\ref{fig:compwise} reports per-component reference trajectories on the same corpus, showing that, for the best molecule in the pre-training set per BO step, the best attainable values for individual objectives also undergo clear transitions aligned with the weight schedule, confirming that the underlying landscape shifts over time in a significant manner.

Importantly, our optimization budget is tiny relative to this corpus: with $T=600$ and $B=10$, we make about $6{,}000$ oracle queries in total—roughly $0.5\%$ of the 1.27M molecules in the pre-training dataset.
Exceeding the top 0.1\% baseline (i.e., outperforming all but roughly 1{,}270 molecules in the pre-training set) represents a particularly strong result.

In Figure~\ref{fig:mainwdataset}, \textsc{TALBO} frequently surpasses the 1\% and 0.1\% dataset baselines across tasks and time windows, indicating that it can identify competitive molecules with a restricted budget.
Moreover, for the \emph{Similarity} task, the optimization trajectory approaches the dataset-derived reference curve, indicating that the generative search can nearly recover the best molecules present in the pre-training corpus.
While this behavior has previously been established in static settings~\citep{maus2022local}, our results recover the same qualitative ``beyond-dataset'' effect in the more challenging drifting-objective regime.
Conversely, when the ``best in dataset'' curve remains far above all methods, the resulting gap highlights the difficulty of reliably reaching rare, well-aligned molecules in a time-varying setting.

\subsection{Ablation studies}
\textbf{Dual Temporal Modeling.}
To disentangle where time-awareness matters, we ablate the two temporal components of \textsc{TALBO}:
(i) a \emph{time-aware surrogate} that models $g(\z,t)$ with a spatio-temporal GP, and
(ii) a \emph{time-aware representation} in which time enters the generative mapping through the covariates (Section~\ref{sec:dgbfgp}).
We compare the full model against three variants:
\emph{(a) no time} (neither component uses $t$),
\emph{(b) surrogate-only} (spatio-temporal GP on $(\z,t)$, but a time-independent representation),
and \emph{(c) representation-only} (time-dependent representation, but a time-agnostic GP on $\z$).
Figure~\ref{fig:dualab} reports baseline ranks over the optimization trajectory under the same budgets as the main benchmark ($\ell_w=0.2$).
Overall, explicitly modeling time improves performance, while the relative impact of time-awareness in the representation versus the surrogate varies across tasks.

\textbf{Sensitivity to priority drift speed.}
Table~\ref{tab:lengthscale} reports the final baseline rank averaged over all tasks for GP lengthscales 
$\ell_{w} \in \{0.1, 0.2, 0.3\}$. 
Here, $\ell_{w}$ is the (relative) lengthscale of the temporal kernel $k_{w}$ used to generate the weight trajectories in Equation~\ref{eq:weight}. 
The effective temporal correlation length is $\ell_w T$, where $T=600$ is the optimization horizon; thus, $\ell_w=0.1,0.2,0.3$ correspond to characteristic drift scales of approximately $60,120,$ and $180$ iterations, respectively. 
Smaller $\ell_{w}$ induces faster priority changes, while larger $\ell_w$ yields smoother, more slowly varying objectives 
Across all tested regimes, \textsc{TALBO} achieves the lowest average rank, indicating consistent superiority under both rapid and gradual priority shifts. 
Full per-task curves are provided in Figure~\ref{fig:appresls}, confirming that these trends are stable.

\textbf{Time-invariant objective.}
A natural concern is that explicitly modeling time could hurt performance when the objective is fixed.
Since one typically cannot know \emph{a priori} whether drift will occur, it is important to verify that the method does not incur a systematic penalty in static settings.
Figure~\ref{fig:constantweight} reports best-found trajectories for three representative tasks under a time-invariant objective.
In this regime, the benefit of time-awareness naturally diminishes.
\textsc{TALBO} remains competitive on some tasks, although it does not consistently outperform the strongest static baselines.
Differences are task-dependent and may reflect architectural variations between the underlying generative models (e.g., DGBFGP versus standard VAEs).
Overall, these results indicate no pronounced degradation when drift is absent.

\section{Discussion}

We introduced \emph{time-varying latent-space Bayesian optimization}, where the objective evolves across rounds while search is conducted in a learned representation.
We proposed \textsc{TALBO}, which incorporates time in both the surrogate and the representation, and showed robust performance across dynamic molecular design tasks, with ablations indicating task-dependent contributions of both temporal components.

\textbf{Limitations and outlook.}
Our benchmarks induce drift through smoothly changing scalarization weights in multi-property optimization.
This isolates a clean form of preference drift, but real campaigns may exhibit richer changes (e.g., updated assays, shifting constraints, abrupt regime changes, or other sources of non-repeatability).
Importantly, our main tracking metric $\tilde f_t=\max_{i\le t} f_t(\x_i)$ involves re-scoring past molecules under the current objective.
As such, $\tilde f_t$ measures \emph{tracking}: whether the optimizer maintains a set of candidates that remains competitive as preferences drift.
In experimental regimes where $f_t$ reflects new measurements or conditions, however, re-scoring past designs may be infeasible, and the instantaneous value $f_t(\x_t)$ becomes the primary observable.
Encouragingly, Figure~\ref{fig:instant} shows \textsc{TALBO} remains competitive under the instantaneous view $f_t(\x_t)$.
Relevant avenues for future work include principled stale-data management in time-varying LSBO.
Dynamic BO has explored forgetting mechanisms such as discounting, sliding windows, resets, and relevance-based pruning to handle observations collected under earlier objective slices that are no longer predictive for the current one~\citep{bogunovic2016time,bardou2024too}.
Extending such policies to LSBO is particularly appealing because staleness can arise not only in the surrogate dataset (outdated labels), but also at the representation level: as the latent geometry evolves through retraining or explicit time-conditioning, the semantics of past latent observations may shift.
Finally, our temporal surrogate relies on a simple stationary time kernel; more expressive temporal models (e.g., nonstationary kernels, change-point kernels, or state-space formulations) could better capture abrupt or multi-scale drift.

\bibliography{refs}

\newpage
\appendix

\setcounter{figure}{0}
\setcounter{table}{0}
\setcounter{equation}{0} 
\renewcommand{\thefigure}{S\arabic{figure}}
\renewcommand{\thetable}{S\arabic{table}}
\renewcommand{\theequation}{S\arabic{equation}}
\renewcommand{\theHequation}{S\arabic{equation}}

\onecolumn

\title{\papertitle\\(Supplementary Material)}
\maketitle

\appendix

\begin{figure}[H]
    \centering
    \includegraphics[width=.95\linewidth]{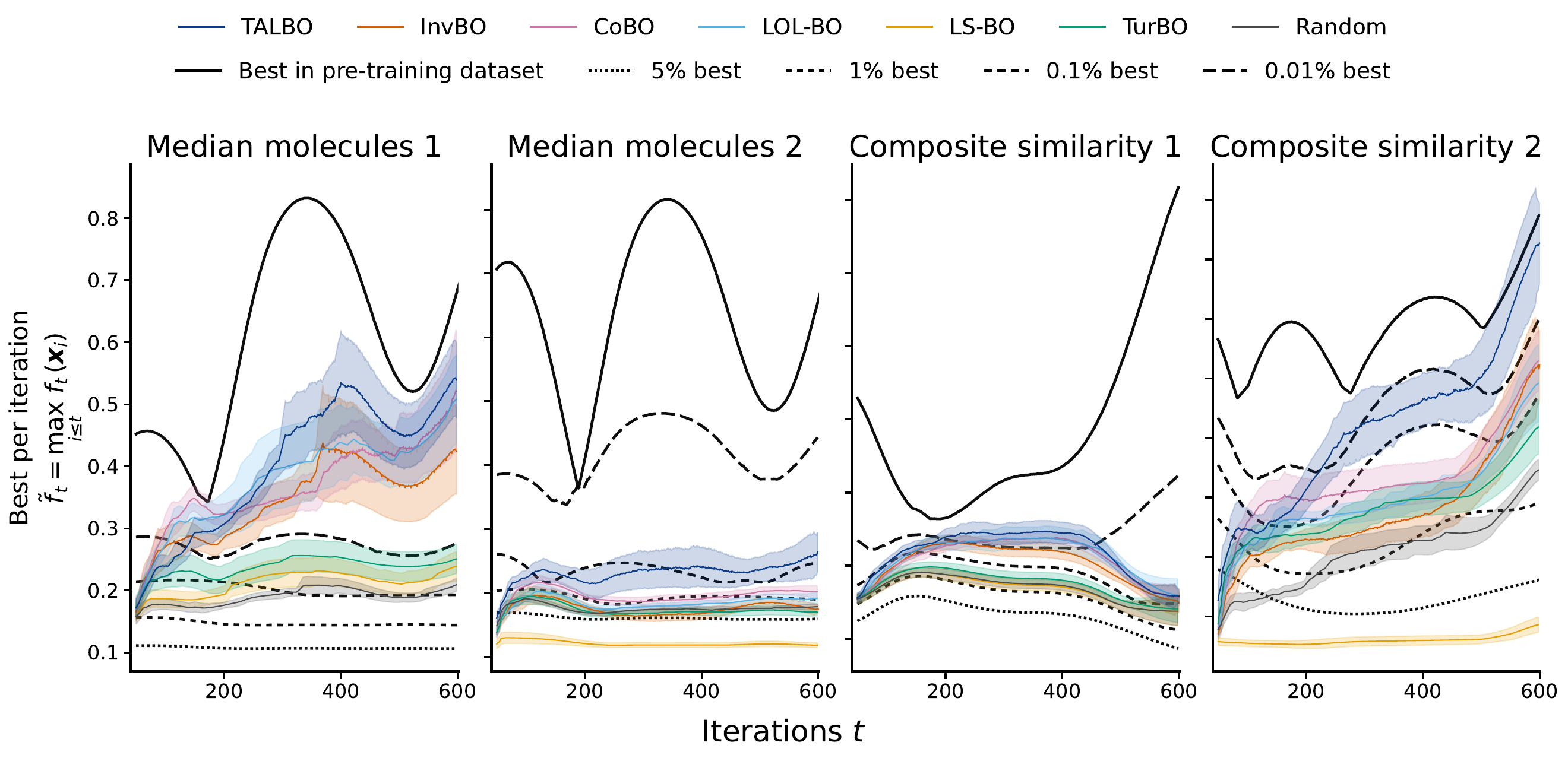}
\caption{
\textbf{Performance comparison with pre-training dataset baselines.}
Current-objective best-so-far value for each dynamic multi-objective task, augmented with statistics derived from the pre-training dataset.
Colored curves correspond to optimization baselines (mean $\pm \tfrac{1}{2}$ std over 10 random seeds).
The solid black curve denotes the best achievable value within the pre-training dataset at each time step, i.e., $\max_j f_t(x_j)$ evaluated over all molecules in the dataset.
Dashed black curves indicate top-percentile subsets of the dataset (5\%, 1\%, 0.1\%, and 0.01\% best), providing reference performance levels attainable without adaptive optimization.
All trajectories are smoothed using a centered moving average for improved readability.
\textbf{The comparison highlights both how challenging it is to outperform strong pre-training set reference levels with only a few thousand queries, and the extent to which dynamic BO methods can nonetheless surpass best-in-dataset metrics.}
}
    \label{fig:mainwdataset}
\end{figure}

\begin{figure}[H]
    \centering
    \includegraphics[width=0.65\linewidth]{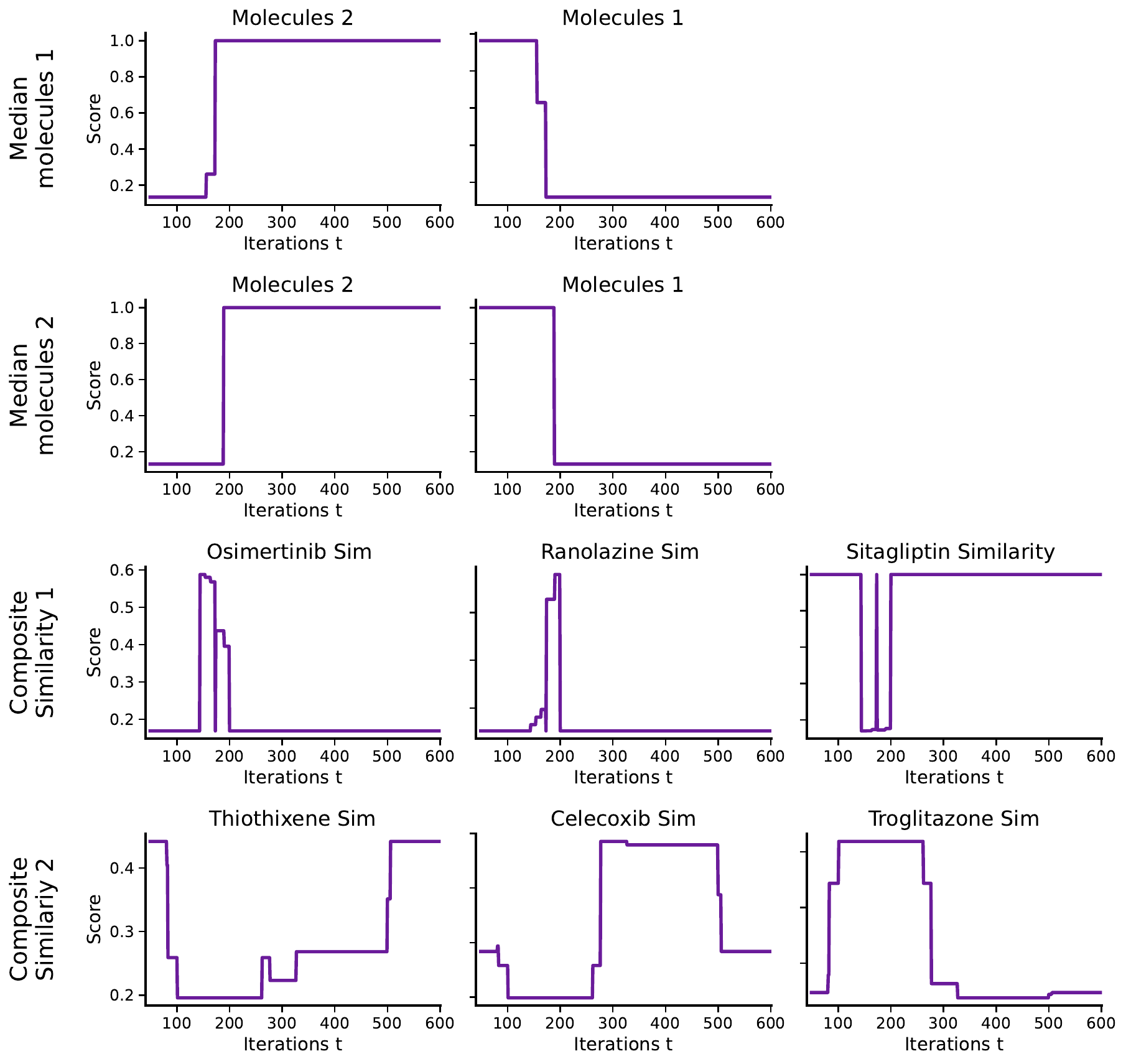}
\caption{
\textbf{Per-component dataset optima under drifting priorities.}
For each task, we evaluate each objective component on the full GuacaMol pre-training corpus (1.27M molecules) using the same time-dependent weighting schedule as in the benchmark, and plot the best attainable component value at each iteration.
\textbf{The resulting trajectories exhibit clear regime changes and transitions across components, confirming that the weight schedule induces meaningful time variation in the underlying objective landscape.}
}
\label{fig:compwise}
\end{figure}

\begin{figure}[H]
    \centering
    \includegraphics[width=0.95\linewidth]{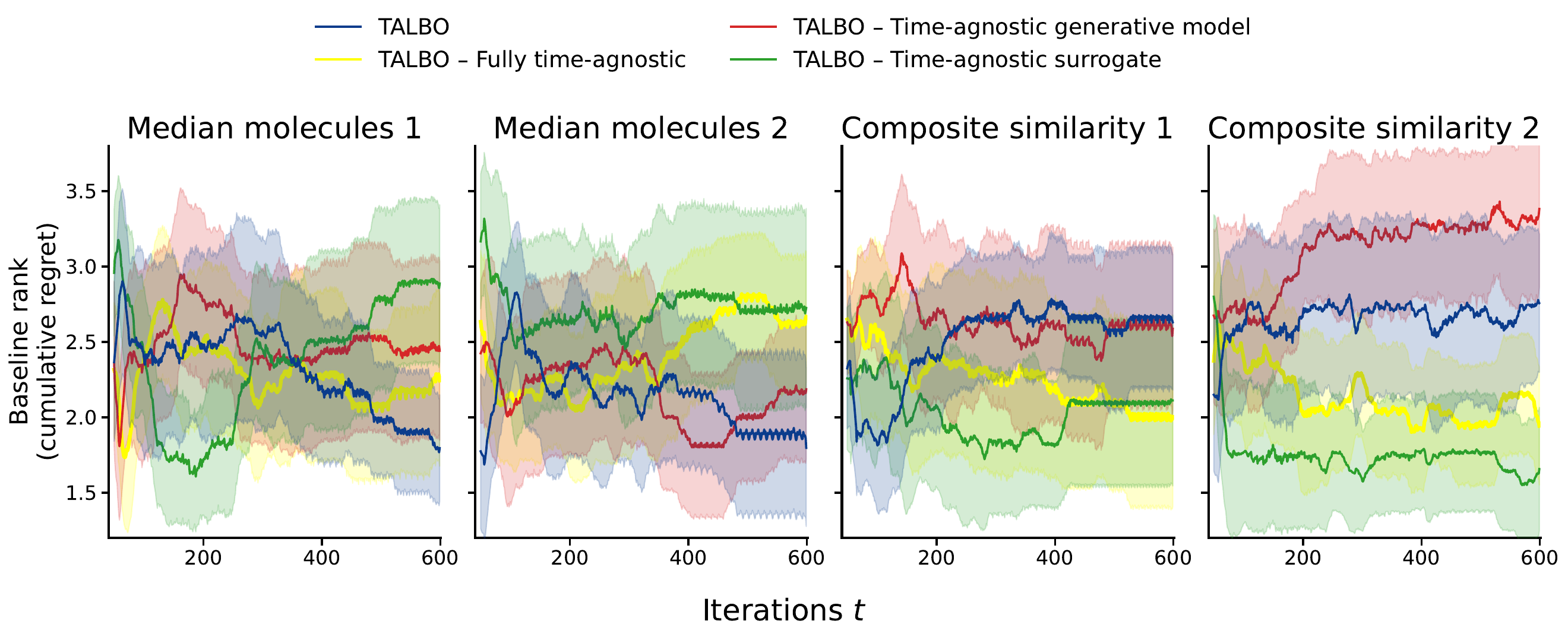}
\caption{
\textbf{Ablation on the dual temporal modeling mechanism.}
For each task and iteration, baselines are ranked within each seed using cumulative regret, and then averaged across 10 random seeds.
Curves show the mean $\pm \tfrac{1}{2}$ standard deviation over seeds. The setting is similar as Figure~\ref{fig:results}: $\ell_w=0.2$.
\textbf{Explicitly modeling time leads to overall performance gains, though the relative contribution of temporal modeling in the generative model versus the surrogate varies across tasks.}
}
\label{fig:dualab}
\end{figure}

\begin{figure}[H]
    \centering
    \includegraphics[width=.75\linewidth]{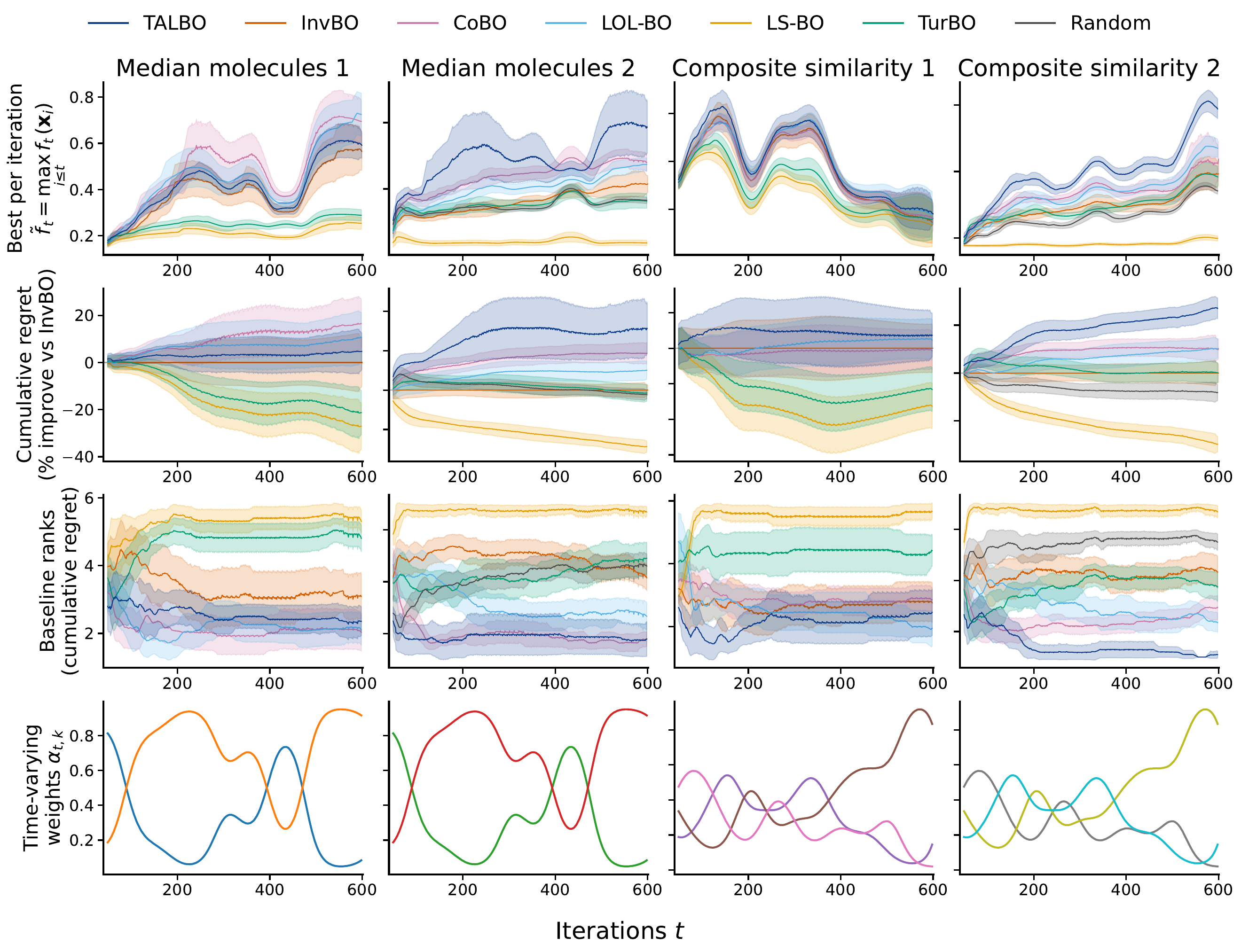}

    
    \includegraphics[width=.75\linewidth]{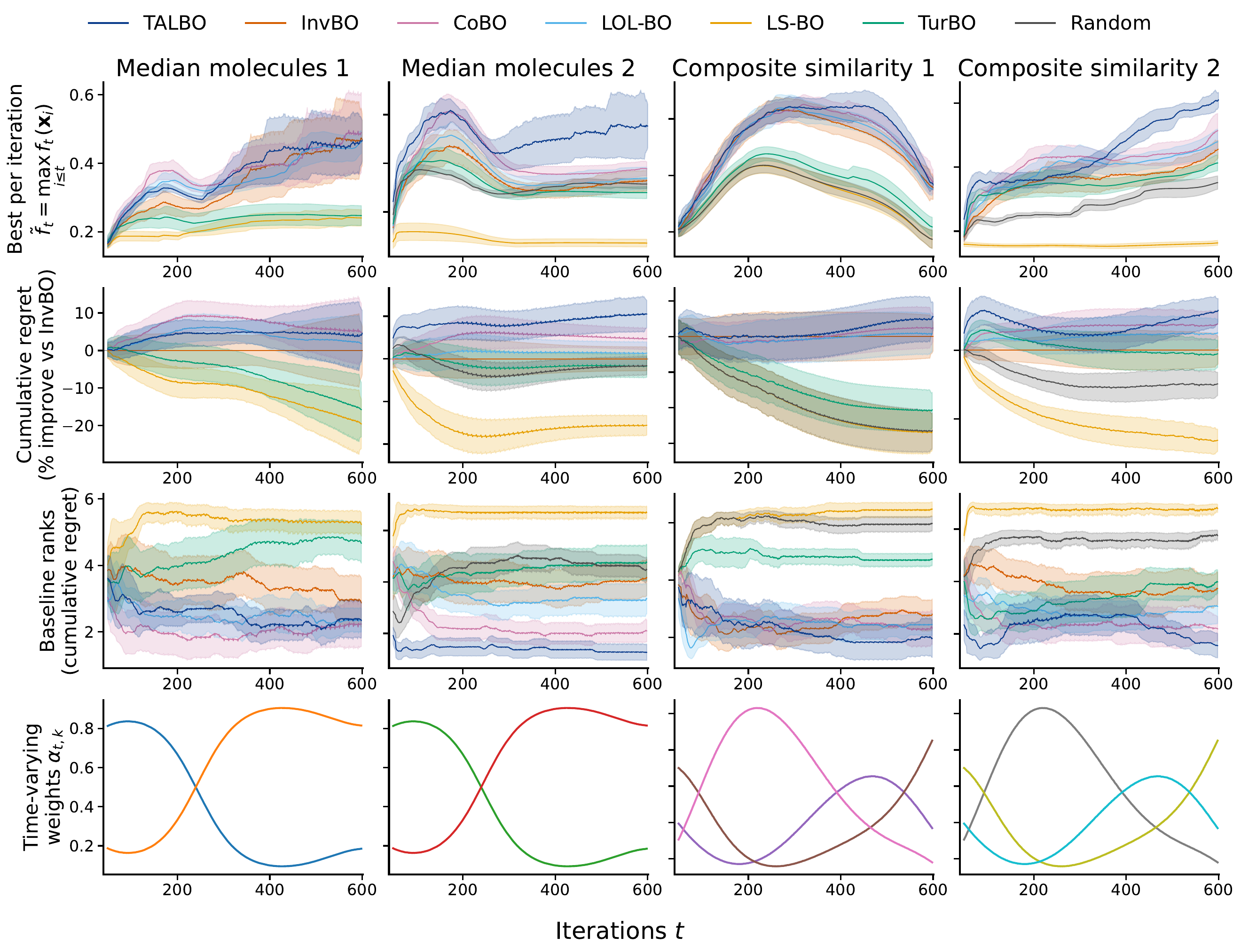}
\caption{
\textbf{Performance comparison across dynamic multi-objective tasks for different GP lengthscales $\ell_w={0.1}$ (top figure) and $\ell_w=0.3$ (bottom figure).}
First row: current-objective best-so-far value.
Second row: cumulative regret, expressed as percentage improvement relative to InvBO.
Third row: baseline rank at each iteration based on cumulative regret (lower is better).
Fourth row: time-varying objective weights governing the dynamic multi-objective trade-off.
Mean $\pm \tfrac{1}{2}$std over 10 random seeds, 
All trajectories are smoothed using a centered moving average for improved readability.
}
    \label{fig:appresls}
\end{figure}

\begin{figure}[H]
    \centering
    \includegraphics[width=.95\linewidth]{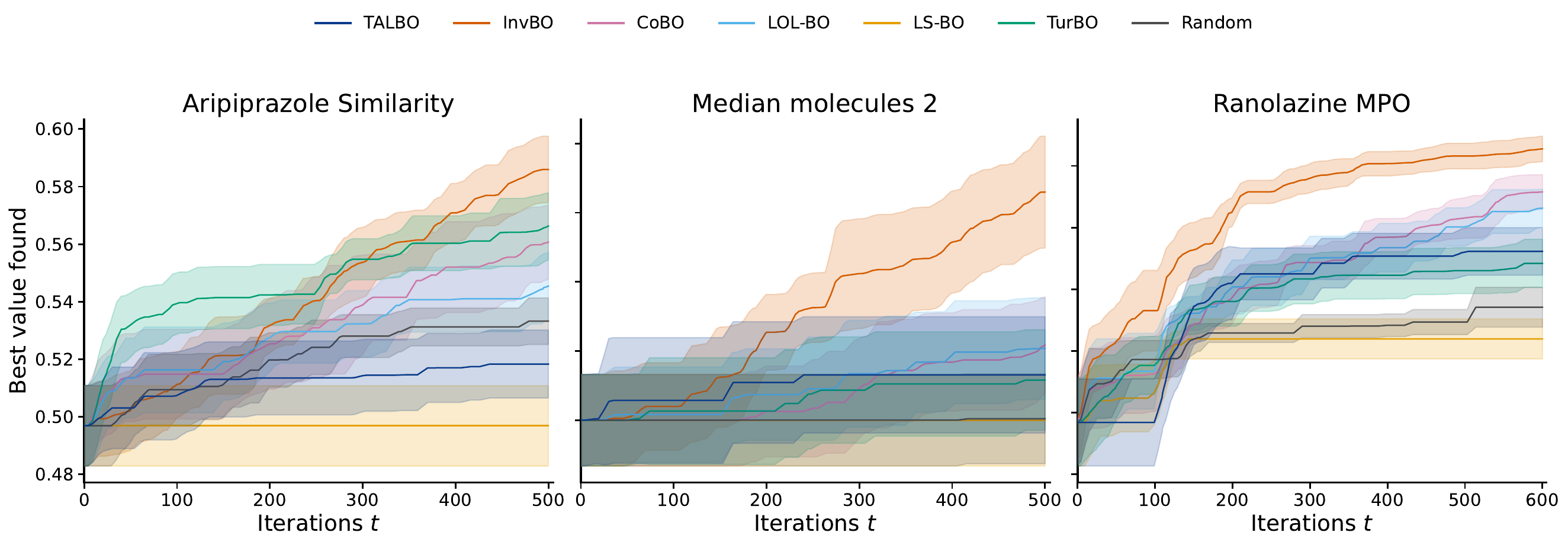}
\caption{
\textbf{Time-invariant objective.}
Best-found value as a function of oracle calls for three representative tasks when the objective remains fixed over time.
Mean $\pm \tfrac{1}{2}$ std over 10 random seeds.
In the absence of temporal drift, differences between methods are task-dependent, and time-awareness does not provide a systematic advantage, as expected.
All trajectories are smoothed using a centered moving average for readability.
\textbf{\textsc{TALBO} remains competitive without exhibiting systematic degradation in the static setting.}
}
    \label{fig:constantweight}
\end{figure}

\begin{figure}[H]
    \centering
    \includegraphics[width=.95\linewidth]{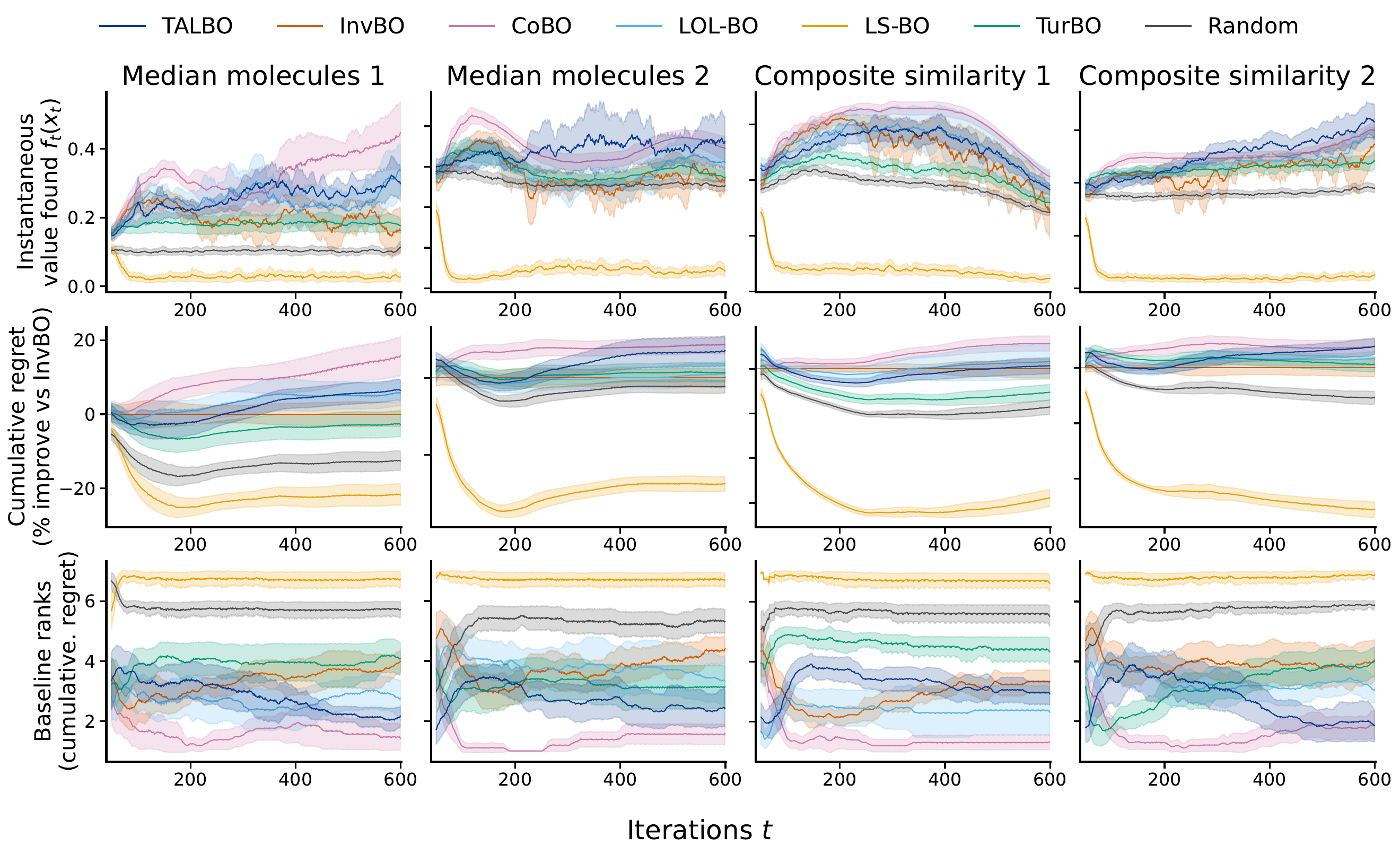}
\caption{
\textbf{Performance comparison across dynamic multi-objective tasks (instantaneous evaluation).}
First row: instantaneous objective value $f_t(\mathbf{x}_t)$ evaluated at each iteration.
Second row: cumulative regret, expressed as percentage improvement relative to \textsc{InvBO}.
Third row: baseline rank at each iteration based on cumulative regret (lower is better).
Mean $\pm \tfrac{1}{2}$std over 10 random seeds.
All trajectories are smoothed using a moving average for improved readability.
\textbf{\textsc{TALBO} consistently achieves competitive instantaneous performance and favorable regret-based ranking across tasks.}
}
    \label{fig:instant}
\end{figure}

\begin{figure}[H]
    \centering
    \includegraphics[width=0.95\linewidth]{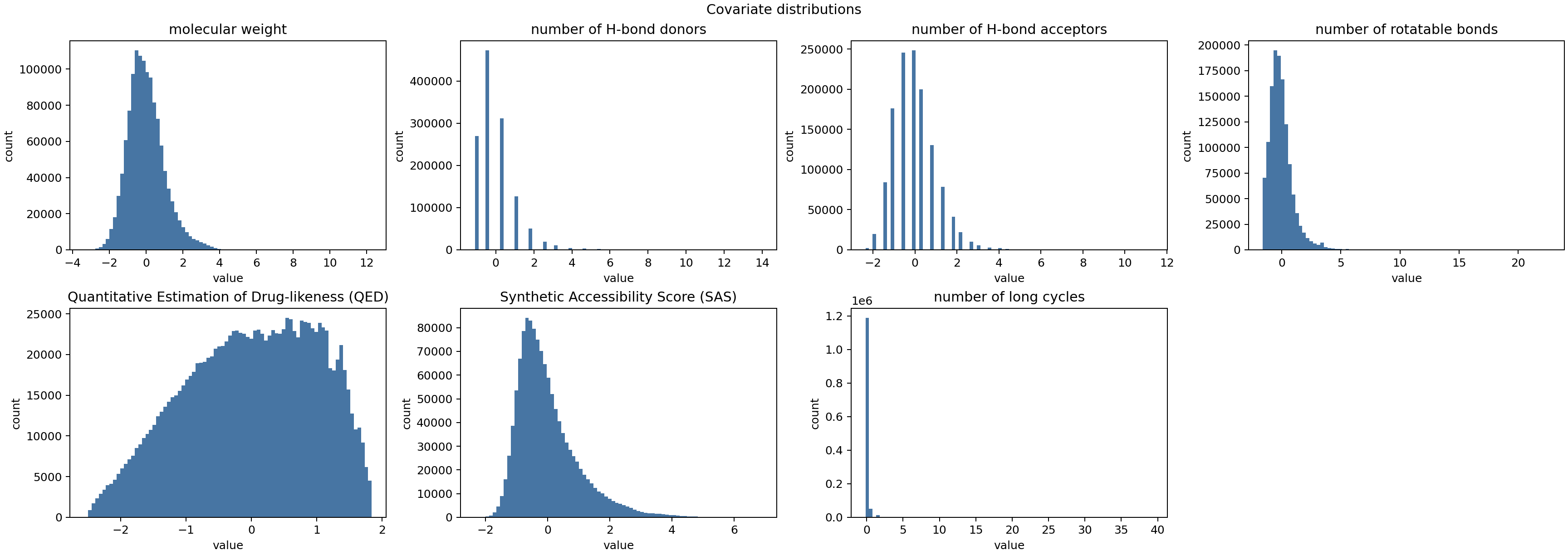}
\caption{
Histograms of molecular covariates used during pretraining (Table~\ref{table:cov}).
}
\label{fig:covariate_distributions}
\end{figure}

The appendix is organized as follows:

\begin{itemize}
    \item Appendix~\ref{app:additional_results} reports complementary experimental results referenced in the main text, including:
    \begin{itemize}[itemsep=1pt, topsep=1pt]
        \item performance comparison with pre-training dataset baselines (Figure~\ref{fig:mainwdataset});
        \item per-component dataset optima under drifting priorities (Figure~\ref{fig:compwise});
        \item ablation on dual temporal modeling (Figure~\ref{fig:dualab});
        \item additional results for different GP lengthscales used to generate drifting weight schedules (Figure~\ref{fig:appresls});
        \item time-invariant objective results (Figure~\ref{fig:constantweight}).
        \item Instantaneous value found metric results (Figure~\ref{fig:instant}).

    \end{itemize}

    \item Appendix~\ref{app:baseline_details} provides implementation details for \textsc{TALBO}'s components and closely related mechanisms, including:
    \begin{itemize}[itemsep=1pt, topsep=1pt]
        \item DGBFGP implementation details (Appendix~\ref{app:dgbfgp});
        \item CoBO-style latent alignment regularizers (Appendix~\ref{app:additional-loss-terms});
        \item InvBO-style latent inversion (Appendix~\ref{app:invbo});
        \item SVGP surrogate objective with deep kernel (Appendix~\ref{app:svgp_loss});
        \item trust-region Bayesian optimization (TuRBO) in latent space (Appendix~\ref{sec:turbo_full});
        \item the comprehensive \textsc{TALBO} loop (Appendix~\ref{app:talbolong});
        \item background on VAEs and GP-prior VAEs (Appendix~\ref{app:vae_background}).
    \end{itemize}

    \item Appendix~\ref{app:expdetails} documents implementation-level details of our experimental setup, including:
    \begin{itemize}[itemsep=1pt, topsep=1pt]
        \item global hyperparameters (Appendix~\ref{app:hyperparameters}, Table~\ref{tab:exp_settings});
        \item DGBFGP configuration and covariates (Appendix~\ref{sec:dgbfgp_full}, Tables~\ref{tab:dgbfgpconfig}--\ref{table:cov} and Figure~\ref{fig:covariate_distributions});
        \item InfoTransformerVAE configuration (Appendix~\ref{sec:infovae_config});
        \item computational resources (Appendix~\ref{app:compute}).
    \end{itemize}
\end{itemize}

\section{Additional results}\label{app:additional_results}

\newpage

\section{Baseline and component details}\label{app:baseline_details}

\subsection{Deep Generative Basis Function GP (DGBFGP)}\label{app:dgbfgp}

This section provides implementation details for the GP-prior latent construction
introduced in Section~\ref{sec:dgbfgp}, including the Hilbert-space approximation
used for scalable GP priors and the decoder likelihood employed in practice. Our presentation follows \citep{balk2025bayesian}. 

\paragraph{Kernel approximation for continuous covariates.}
For a univariate continuous covariate $c \in\mathbb{R}$,
we use the squared-exponential (SE) kernel for component $r$ (index $r$ is suppressed from $c^{(r)}$ for clarity)
\begin{align}
k_{\mathrm{se}}^{(r)}\!\big(c,c^\prime \mid \sigma_r,\ell_r\big)
= \sigma_r^2
\exp\!\left(
-\frac{\big(c-c^{\prime}\big)^2}{2\ell_r^2}
\right).
\end{align}
Stationary kernels admit a spectral density $s(\omega)$ via the Wiener--Khintchin theorem,
\begin{align}
s(\omega)
&= \int_{-\infty}^{\infty} k(u)\,e^{-i\omega u}\,du, \\
k(u)
&= \frac{1}{2\pi}\int_{-\infty}^{\infty} s(\omega)\,e^{i\omega u}\,d\omega, \notag
\end{align}
and for the SE kernel the spectral density is
\begin{align}
s_{\mathrm{se}}(\omega \mid \sigma_r,\ell_r)
= \sigma_r^2 \ell_r \sqrt{2\pi}\,\exp\!\left(-\tfrac{1}{2}\ell_r^2 \omega^2\right).
\end{align}

Using the Hilbert-space (eigenfunction) approximation on a bounded domain
$\Omega=[-J,J]$ with Dirichlet boundary conditions,
\begin{align}
k_{\mathrm{se}}^{(r)}\!\big(c,c^{\prime}\big)
\approx 
\tilde{k}^{(r)}\!\big(c,c^{\prime}\big)
= \sum_{m=1}^{M}
s_{\mathrm{se}}(\sqrt{\lambda_m}\mid\sigma_r,\ell_r)\,
\phi_m^{(r)}(c)\,\phi_m^{(r)}(c^{\prime}),
\end{align}
where the Laplacian eigenpairs are
\begin{align}
\phi_m(c)
&= \frac{1}{\sqrt{J}}\sin\!\left(\frac{\pi m (c + J)}{2J}\right), \\
\lambda_m
&= \left(\frac{\pi m}{2J}\right)^2. \notag
\end{align}
Note that we have dropped index $r$ from $\phi_m(c)$ and $\lambda_m$ for clarity because they depend on $r$ only via the size of the domain $\Omega = [-J,J]$ of the $r$th covariate component. 
Defining $\boldsymbol{\phi}^{(r)}(c)=[\phi_1(c),\ldots,\phi_M(c)]^\top$,
a GP draw for latent dimension $l$ can be written in linear form as
\begin{align}
h^{(r,l)}(c)
&\approx \boldsymbol{a}^{(r,l)\top}\boldsymbol{\phi}^{(r)}(c),
\qquad
\boldsymbol{a}^{(r,l)} \sim
\mathcal{N}\!\bigl(\mathbf{0}, \boldsymbol{S}^{(r)}(\sigma_r,\ell_r)\bigr),
\end{align}
with diagonal covariance
\begin{align}
\boldsymbol{S}^{(r)}(\sigma_r,\ell_r)
=
\operatorname{diag}\!\bigl(
s_{\mathrm{se}}(\sqrt{\lambda_1}\mid\sigma_r,\ell_r),\ldots,
s_{\mathrm{se}}(\sqrt{\lambda_M}\mid\sigma_r,\ell_r)
\bigr).
\end{align}
In our implementation we use Hilbert features with $J=2.55$ and $M=128$.

\paragraph{Decoder likelihood (categorical tokens).}
For structured discrete designs (e.g., SELFIES strings),
the decoder $p_\theta(\x\mid\z)$ factorizes autoregressively over tokens.
Given a token sequence $\x=(x_1,\dots,x_L)$ and prefix $x_{<\ell}$,
the decoder defines per-step categorical probabilities
\begin{align}
\pi_\ell(\z,x_{<\ell})
:= p(x_{\ell} \mid \z, x_{< \ell}) = \mathrm{softmax}\!\big(\mathrm{logits}_\ell(\z,x_{<\ell})\big),
\qquad \ell=1,\dots,L.
\end{align}
The reconstruction term in Equation~\ref{eq:finalloss} is implemented
as the length-normalized decoder log-likelihood
\begin{align}
\log p_\theta(\x \mid \z)
=
\frac{1}{L}\sum_{\ell=1}^{L}
\log \pi_\ell(\z,x_{<\ell})_{x_\ell}.
\end{align}

\subsection{CoBO-style latent alignment regularizers}
\label{app:additional-loss-terms}

When fine-tuning the generative representation during BO (Algorithm~\ref{alg:tvlsbo_dgbfgp}),
optimizing reconstruction alone does not guarantee that distances in latent space reflect changes in the black-box
objective. This ``geometry--objective'' mismatch is a known failure mode in LSBO, and CoBO~\citep{lee2023advancing}
proposed simple regularizers to encourage local smoothness of the objective as a function of the latent code, and a non-degenerate latent scale. \textsc{TALBO} adopts the same regularizers and weighting schemes, which we now describe briefly.

\paragraph{Objective-based importance weights.}
To focus regularization on promising designs, we assign each example a soft importance weight
$w_i \in (0,1]$ derived from its objective value.
Following~\citet{lee2023advancing}, let $y_q$ be the $q$-quantile of the objective values in the current minibatch
(and $\sigma_w>0$ a smoothing parameter). Define
\begin{equation}
w_i \;:=\; \lambda(y_i)
\;=\; \Pr(Y > y_q),
\qquad Y \sim \mathcal{N}(y_i,\sigma_w^2),
\label{eq:cobo-weight}
\end{equation}
equivalently $w_i = 1-\Phi\!\big((y_q-y_i)/\sigma_w\big)$ where $\Phi$ is the standard Gaussian CDF.

\paragraph{Weighted Lipschitz regularizer.}
CoBO motivates smoothness by encouraging the composite objective in latent space to behave as a Lipschitz function.
Concretely, for a minibatch $\{(\z_i,y_i,w_i)\}_{i=1}^N$, define pairwise slopes
\begin{equation}
s_{ij} := \frac{|y_i-y_j|}{\|\z_i-\z_j\|_2},\qquad
L := \mathrm{median}\big(\{s_{ij}\}_{i,j}\big),
\label{eq:cobo-slopes}
\end{equation}
and penalize only slopes that exceed the typical (median) value:
\begin{equation}
\mathcal{L}_{\mathrm{Lip}}^{(w)}
:= \frac{1}{N^2}\sum_{i=1}^N\sum_{j=1}^N
\sqrt{w_i w_j}\;
\max\!\big(0,\, s_{ij}-L\big).
\label{eq:cobo-lip}
\end{equation}
The geometric-mean weighting $\sqrt{w_iw_j}$ emphasizes smoothness in high-value regions, as proposed in CoBO.

\paragraph{Latent-space scale regularizer.}
A caveat is that $\mathcal{L}_{\mathrm{Lip}}^{(w)}$ alone admits degenerate solutions (e.g., inflating all pairwise
distances $\|\z_i-\z_j\|_2$ shrinks $s_{ij}$). CoBO therefore constrains the \emph{average} latent scale.
Let
\begin{equation}
\bar d_{\z} := \frac{1}{N^2}\sum_{i=1}^N\sum_{j=1}^N \|\z_i-\z_j\|_2,
\qquad
c_d := \mathbb{E}\big[\|U-V\|_2\big]
= 2\,\frac{\Gamma\!\big(\frac{d+1}{2}\big)}{\Gamma\!\big(\frac{d}{2}\big)},
\label{eq:cobo-scale}
\end{equation}
where $U,V\sim \mathcal{N}(0,I_d)$ and $d$ is the latent dimension.
We then define
\begin{equation}
\mathcal{L}_Z := \big|\bar d_z - c_d\big|.
\label{eq:cobo-lz}
\end{equation}
This anchors the latent geometry to the scale implied by a standard normal prior, preventing trivial rescalings.

\paragraph{Full objective during representation updates.}
We update the generative model during BO steps by augmenting the DGBFGP objective with these alignment terms, and (when jointly
training with the surrogate) also include the surrogate loss $\mathcal{L}_{\mathrm{SVGP}}$ (Appendix~\ref{app:svgp_loss}):
\begin{equation}
\mathcal{L}_{\textsc{TALBO}}
= - \mathcal{L}_{\textsc{DGBFGP}}
+ \mathcal{L}_{\mathrm{Lip}}^{(w)}
+ \mathcal{L}_Z
+ \mathcal{L}_{\mathrm{SVGP}}.
\label{eq:talbo-full-loss}
\end{equation}

\subsection{INVBO-Style Latent inversion}\label{app:invbo}
A recurring failure mode in LSBO is \emph{misalignment} between the latent codes used to train the surrogate and the discrete designs actually evaluated by the oracle.
Concretely, if a latent code $\z$ associated with an evaluated design $\x$ does not decode back to $\x$ under the current decoder (i.e., $\hat \x=\mathrm{Dec}_\theta(\z)\neq \x$), then the training pair $(\z,y)$ with $y=f_t(\x)$ no longer corresponds to the surrogate target $f_t(\mathrm{Dec}_\theta(\z))$.
This mismatch can be exacerbated after representation/decoder updates.
Following the plug-and-play \emph{inversion} module of InvBO~\citep{chu2024inversion}, we optionally refine latent codes by explicitly \emph{inverting} the decoder to recover a code that reconstructs the evaluated design, without requiring additional oracle calls.

Given a design $\x\in\mathcal{X}$, we solve
\begin{align}
\z^{\mathrm{inv}}(\x)
&\in \arg\min_{\z\in\mathbb{R}^d}\; \mathcal{L}_{\mathrm{inv}}(\z;\x), \label{eq:inv_def}\\
\mathcal{L}_{\mathrm{inv}}(\z;\x)
&:= -\log p_\theta(\x \mid \z), \label{eq:inv_loss}
\end{align}
where $-\log p_\theta(\x \mid \z)$ measures how well the decoder reconstructs $\x$ from $\z$.
In our SELFIES setting, $\mathcal{L}_{\mathrm{inv}}(\z;\x)$ is implemented as the (length-normalized) token negative log-likelihood under the decoder~\citep{maus2022local}.

We initialize from the current encoder/embedding output (e.g., $\z^{(0)}\sim q_\phi(\z\mid \x)$, or $\z^{(0)}=\mathbb{E}_{q}[\z(\mathbf{c}(\x,t);\boldsymbol{A})]$) and update only $\z$ while keeping decoder parameters fixed:
\begin{equation}
\z^{(k+1)} \leftarrow \z^{(k)} - \eta\,\nabla_{\z} \mathcal{L}_{\mathrm{inv}}(\z^{(k)};\x),
\qquad k=0,\dots,T_{\mathrm{inv}}-1.
\label{eq:inv_update}
\end{equation}
As in InvBO, we may stop early once the decoded molecule $\hat \x=\Gamma(\z^{(k+1)})$ is sufficiently close to $\x$ under an input-space distance $d_{\mathcal{X}}(\cdot,\cdot)$ (e.g., normalized Levenshtein distance on SELFIES strings),
\begin{equation}
d_{\mathcal{X}}(\x,\hat \x)\le \epsilon.
\label{eq:inv_stop}
\end{equation}
The refined latent $\z^{\mathrm{inv}}(\x)$ can then replace stored latent codes for newly added points and/or the current top-$k$ set, maintaining an \emph{aligned} dataset for surrogate learning under decoder updates.

\subsection{SVGP surrogate objective with deep kernel}\label{app:svgp_loss}

At iteration $t$, the surrogate models the latent objective
\[
g(\z,t)\;:=\;f_t(\Gamma(\z)),
\]
using the dataset $\mathcal{D}_t=\{(\z_i,t_i,y_i)\}_{i=1}^{n_t}$.
We place a GP prior on $g$,
\[
g \sim \mathcal{GP}\!\big(0,\,k((\z,t),(\z',t'))\big),
\qquad
k((\z,t),(\z',t')) = k_{\z}(\z,\z')\,k_t(t,t'),
\]
and use the Gaussian likelihood
\[
p(y_i \mid g(\z_i,t_i)) = \mathcal{N}\!\big(y_i \mid g(\z_i,t_i), \sigma_y^2\big).
\]

To scale beyond $n_t\gg 10^3$, we use a sparse variational GP (SVGP)~\citep{hensman2015scalable}
with inducing inputs $\mathbf{U}=\{u_m\}_{m=1}^{M}$ in the joint input space (e.g. $u_m=(\z_m,t_m)$),
and variational distribution
\[
q(\mathbf{g}_{\mathbf{U}})=\mathcal{N}(\mathbf{m},\mathbf{S}),
\]
where $\mathbf{g}_{\mathbf{U}} := \big(g(u_1),\ldots,g(u_M)\big)$.
Let $p(\mathbf{g}_{\mathbf{U}})=\mathcal{N}(\mathbf{0},\mathbf{K}_{\mathbf{UU}})$ denote the GP prior
at inducing inputs. The SVGP ELBO is
\begin{equation}
\mathcal{L}_{\mathrm{SVGP}}
=
\sum_{i=1}^{n_t}\mathbb{E}_{q(g(\z_i,t_i))}\!\big[\log p(y_i \mid g(\z_i,t_i))\big]
-
\mathrm{KL}\!\big(q(\mathbf{g}_{\mathbf{U}})\,\|\,p(\mathbf{g}_{\mathbf{U}})\big),
\label{eq:svgp_elbo}
\end{equation}
and is optimized w.r.t.\ variational parameters $(\mathbf{m},\mathbf{S})$ and kernel hyperparameters
(typically with minibatching over $i$).

\paragraph{Deep kernel learning.}
When using a deep kernel~\citep{wilson2016deep}, we replace $\z$ by a learned feature map
$\phi_\psi(\z)$ and define
\[
k_{\z}(\z,\z') \;:=\; k_{\mathrm{RBF}}\!\big(\phi_\psi(\z),\phi_\psi(\z')\big),
\]
so that the surrogate input becomes $(\phi_\psi(\z),t)$. We then optimize $\psi$ jointly with
the SVGP objective in Equation~\ref{eq:svgp_elbo}.

\subsection{Trust region Bayesian Optimization configuration}\label{sec:turbo_full}
TurBO-M~\citep{eriksson2019scalable} is a widely adopted strategy for high-dimensional BO that operates by maintaining $M$ independent local searches, each confined within its own hyper-rectangular trust region. TurBO-1 was shown to outperform other variants in most settings and was subsequently adopted for LSBO in \cite{maus2022local}. Although LSBO maps the original input space $\mathcal{X}$ to a compressed latent space $\mathcal{Z}$, the resulting search space remains high-dimensional, posing challenges for traditional BO approaches. TurBO is used to improve search efficiency by centering the optimization process at the current incumbent and creating a trust region around it, as described below:

A trust region centered at $\boldsymbol{\kappa}\in\mathbb{R}^d$ with length $L>0$ and weight $\boldsymbol{\omega} \in \mathbb{R}^d$ is
\begin{align}
\mathcal{T}(\boldsymbol{\kappa};L,\boldsymbol{\omega})
&= \big\{ \z\in\mathbb{R}^d : \boldsymbol{\kappa} - \tfrac{1}{2} \boldsymbol{\omega} L \le \z \le \boldsymbol{\kappa} + \tfrac{1}{2} \boldsymbol{\omega} L \big\}.
\end{align}

State update for a new batch $Y_{\mathrm{next}}$ (with best value $y^\star$ so far):
\begin{align}
\text{Success if} \quad 
\max\!\left(Y_{\mathrm{next}}\right)
&> y^\star + 10^{-3}\lvert y^\star \rvert,
\quad \text{failure otherwise}, \notag \\
y^\star 
&\leftarrow 
\max\!\Big( y^\star,\; \max\!\left(Y_{\mathrm{next}}\right) \Big).
\end{align}

Let $n_{\mathrm{succ}}$ and $n_{\mathrm{fail}}$ denote the success and failure counters.

On success:
\begin{equation*}
n_{\mathrm{succ}} \leftarrow n_{\mathrm{succ}} + 1,
\qquad
n_{\mathrm{fail}} \leftarrow 0.
\end{equation*}
On failure:
\begin{equation*}
n_{\mathrm{succ}} \leftarrow 0,
\qquad
n_{\mathrm{fail}} \leftarrow n_{\mathrm{fail}} + 1.
\end{equation*}
If $n_{\mathrm{succ}}$ reaches the tolerance $N_{\mathrm{succ}}$, expand $L \leftarrow \min(2L, L_{\max})$, if $n_{\mathrm{fail}}$ reaches the tolerance $N_{\mathrm{fail}}$, shrink $L \leftarrow \frac{L}{2}.$

A restart is triggered if $L < L_{\min}$.

\vspace{-.4cm}

\subsection{Comprehensive \textsc{TALBO} loop}\label{app:talbolong}

\begin{algorithm}[H]
\caption{\textsc{TALBO}, comprehensive version with CoBO-style alignment, and InvBO-style inversion}
\label{alg:talbolong}
\begin{algorithmic}[1]
\REQUIRE Failure tolerance $N_{\mathrm{fail}}$;  improvement threshold $\delta$; initial evaluated set $\mathcal{D}_0=\{(\x_i,t_i,y_i)\}_{i=1}^{n_0}$.
\STATE Pre-train DGBFGP to obtain $q_0(\boldsymbol{A},\sigma,\ell)$ and decoder parameters $\theta_0$.
\STATE $\mathcal{D}_0^{\mathrm{rich}} \gets \emptyset$.
\FOR{each $(\x_i,t_i,y_i)\in\mathcal{D}_0$}
\STATE $\z_i \gets \textsc{EmbedAndAlign}(\x_i,t_i;\,q_0,\theta_0)$
\STATE $\mathcal{D}_0^{\mathrm{rich}} \gets \mathcal{D}_0^{\mathrm{rich}} \cup \{(\x_i,\z_i,t_i,y_i)\}$
\ENDFOR
\STATE $y^\star \gets \max\{y : (\x,\z,t,y)\in \mathcal{D}_0^{\mathrm{rich}}\}$; \quad $n_{\mathrm{fail}} \gets 0$.

\FOR{$t = 1$ \TO $T$ or until criterion satisfied}
    \STATE Fit spatio-temporal surrogate $g_t$ on $\mathcal{D}_{t-1}^{(z)} := \{(\z,t,y):(\x,\z,t,y)\in \mathcal{D}_{t-1}^{\mathrm{rich}}\}$ (Equation~\ref{eq:svgp_elbo})
    \STATE Select $\hat{\z}_t \in \arg\max_{\z\in \mathcal{T}_t} \alpha\!\big(\z;\,p(g_t(\cdot,t)\mid \mathcal{D}_{t-1}^{(z)})\big)$, where $\mathcal{T}_t$ is the current TuRBO trust region centered at the incumbent in latent space (Appendix~\ref{sec:turbo_full}).
    \STATE Decode and evaluate $\x_t \sim p_{\theta_{t-1}}(\cdot\mid \hat{\z}_t)$.
    \STATE $y_t=f_t(\x_t)+\varepsilon_t$.
\STATE $\z_t \gets \textsc{EmbedAndAlign}(\x_t,t;\,q_{t-1},\theta_{t-1})$
\STATE $\mathcal{D}_t^{\mathrm{rich}} \gets 
\mathcal{D}_{t-1}^{\mathrm{rich}} \cup \{(\x_t,\z_t,t,y_t)\}$

    \IF{$y_t > y^\star + \delta\,|y^\star|$}
        \STATE $y^\star \gets y_t$; \quad $n_{\mathrm{fail}} \gets 0$.
    \ELSE
        \STATE $n_{\mathrm{fail}} \gets n_{\mathrm{fail}} + 1$.
    \ENDIF

    \IF{$n_{\mathrm{fail}} \ge N_{\mathrm{fail}}$}
        \STATE Update DGBFGP posterior to $q_t(\boldsymbol{A},\sigma,\ell)$ and decoder to $\theta_t$
        by optimizing Equation~\ref{eq:talbo-full-loss}.
        \STATE $\mathcal{D}_{t,\mathrm{new}}^{\mathrm{rich}} \gets \emptyset$.
        \FOR{each $(\x_i,\z_i,t_i,y_i)\in\mathcal{D}_t^{\mathrm{rich}}$}
\STATE $\z_i \gets \textsc{EmbedAndAlign}(\x_i,t_i;\,q_t,\theta_t)$
\STATE $\mathcal{D}_{t,\mathrm{new}}^{\mathrm{rich}} \gets 
\mathcal{D}_{t,\mathrm{new}}^{\mathrm{rich}} \cup \{(\x_i,\z_i,t_i,y_i)\}$
        \ENDFOR
        \STATE $\mathcal{D}_t^{\mathrm{rich}} \gets \mathcal{D}_{t,\mathrm{new}}^{\mathrm{rich}}$; \quad $n_{\mathrm{fail}} \gets 0$.
    \ELSE
        \STATE $q_t \gets q_{t-1}$; \quad $\theta_t \gets \theta_{t-1}$.
    \ENDIF
\ENDFOR
\vspace{0.5em}
\hrule
\vspace{0.5em}
\STATE \textbf{Procedure} $\textsc{EmbedAndAlign}(\x,t;\,q,\theta)$:
\STATE \hspace{1em} $\mathbf{c} \gets \mathbf{c}(\x,t)$ \texttt{// covariates may include pre-computed descriptors or metadata}
\STATE \hspace{1em} $\z \gets \mathbb{E}_{q}\!\big[\z(\mathbf{c};\boldsymbol{A})\big]$
\STATE \hspace{1em} \texttt{// Inversion step: refine $\z$ to decode back to $\x$ (Section~\ref{app:invbo})} 
\STATE \hspace{1em} $\z^{(0)} \gets \z$
\FOR{$k=1$ \TO $T_{\mathrm{inv}}$}
    \STATE \hspace{2em} $\z^{(k)} \gets \z^{(k-1)} - \eta \nabla_{\z}\mathcal{L}_{\mathrm{inv}}(\z^{(k-1)};\x)$
    \IF{$d_{\mathcal{X}}(\x,\Gamma(\z^{(k)})) \le \epsilon$}
        \STATE \hspace{2em} \textbf{break}
    \ENDIF
\ENDFOR
\STATE \hspace{1em} \textbf{return} $\z^{(k)}$
\end{algorithmic}
\end{algorithm}

\subsection{Background on VAEs and GP-prior VAEs}\label{app:vae_background}

\paragraph{Motivation.}
Latent-space Bayesian optimization typically relies on a pretrained generative model to map structured designs $\x\in\mathcal{X}$ to a continuous latent space $\z\in\mathbb{R}^d$ and decode candidate latents back to designs.
The most common backbone is a VAE~\citep{kingma2013auto}, whose latent prior encourages a well-behaved embedding.
Our method is closely related in spirit, but differs in the \emph{prior}: instead of a simple independent prior (e.g., $\mathcal{N}(\mathbf{0},\mathbf{I})$), we endow the latent representation with a \emph{covariate-dependent GP prior} to obtain a temporally coherent latent geometry.
Here, we briefly review VAEs, conditional VAEs, and GP-prior VAEs as background.

\paragraph{Variational autoencoders.}
A VAE is a latent-variable generative model for an observation $\x\in\mathcal{X}$ of the form
\begin{equation}
p_\omega(\x,\z)=p_\theta(\x\mid \z)\,p(\z), \qquad \omega=\{\theta\},
\end{equation}
where $\z\in\mathbb{R}^d$ is an unobserved latent code and $p(\z)$ is typically a simple prior such as $\mathcal{N}(\mathbf{0},\mathbf{I})$.
The posterior $p_\omega(\z\mid \x)$ is generally intractable, so VAEs use amortized variational inference with an encoder
$q_\phi(\z\mid \x)$ to approximate it~\citep{kingma2013auto,murphy2023probabilistic}.
Learning proceeds by maximizing the evidence lower bound (ELBO):
\begin{equation}
\log p_\omega(\x)
\ge
\mathbb{E}_{q_\phi(\z\mid \x)}\!\left[\log p_\theta(\x\mid \z)\right]
-
\mathrm{KL}\!\left(q_\phi(\z\mid \x)\,\|\,p(\z)\right),
\end{equation}
with respect to encoder and decoder parameters $(\phi,\theta)$.

\paragraph{Conditional VAEs.}
A conditional VAE (cVAE) incorporates auxiliary covariates $\mathbf{c}\in\mathbb{R}^R$ by conditioning both encoder and decoder,
\begin{equation}
p_\omega(\x,\z\mid \mathbf{c}) = p_\theta(\x\mid \z,\mathbf{c})\,p(\z\mid \mathbf{c}),
\qquad
q_\phi(\z\mid \x,\mathbf{c}),
\end{equation}
leading to an ELBO of the same form with $\mathbf{c}$ treated as observed input~\citep{sohn2015learning}.
In our setting, covariates are denoted $\mathbf{c}(\x,t)$ and can include time. Alternative parametrization of the generative model assumes dependency on the covariates $\mathbf{c}$ only in the prior, but not in the decoder, resulting in $p_\omega(\x,\z\mid \mathbf{c}) = p_\theta(\x\mid \z)\,p(\z\mid \mathbf{c})$. 

\paragraph{Gaussian process prior VAEs.}
A limitation of standard VAEs (and cVAEs) in LSBO is that the latent space is learned primarily to reconstruct $\x$ under an unsupervised objective, and the resulting geometry may be poorly aligned with downstream optimization objectives.
GP-prior VAEs replace the simple latent prior with a Gaussian process prior that ties latent codes across examples through shared covariates.
Concretely, letting $\mathbf{c}_n$ denote covariates for $\x_n$, GP-prior VAEs posit latent functions $h^{(j)}(\cdot)$ for each latent dimension $j=1,\dots,d$,
\begin{equation}
h^{(j)}(\cdot)\sim \mathcal{GP}(0,k(\cdot,\cdot)), 
\qquad
\z_n^{(j)} := h^{(j)}(\mathbf{c}_n),
\end{equation}
so that the collection $\{\z_n\}_{n=1}^N$ is no longer independent a priori but varies smoothly with $\mathbf{c}$.
This induces a covariate-aware latent geometry (e.g., longitudinal structure when $\mathbf{c}$ includes time), and has been explored for improving BO in latent spaces by conditioning the GP prior on auxiliary variables~\citep{ramchandran2025highdimensional}.
Our \textsc{TALBO} model that utilizes DGBFGP as its generative model can be viewed as a scalable and time-varying variant of this idea: it uses a basis-function approximation with global variational parameters to make GP-prior latent representations practical at scale.

\section{Experiment setup and implementation details}\label{app:expdetails}
\subsection{Global hyperparameters}\label{app:hyperparameters}
\begin{table}[H]
\centering
\begin{tabular}{ll}
\toprule
\multicolumn{2}{l}{\textbf{Optimization setup}} \\
\midrule
Optimization horizon $T$ & 600 \\
Initial dataset size $N_{\text{init}}$ & 100 \\
Batch size $B$ & 10 \\
Number of seeds & 10 \\
\addlinespace
\multicolumn{2}{l}{\textbf{Time-varying objective generation}} \\
\midrule
Temporal kernel (Equation~\ref{eq:weight}) & Squared-exponential $k_{\text{w}}$ \\
Relative lengthscale $\ell_w$ & $0.2$ (unless otherwise stated) \\
Effective drift scale & $\ell_w T$\\
Temperature $\beta$ (Equation~\ref{eq:softmax}) & 1.0\\
Noise $\sigma^2$ & 0.0\\
\addlinespace
\multicolumn{2}{l}{\textbf{Bayesian optimization configuration}} \\
\midrule
Acquisition function & $q$-Batch Thompson Sampling \\
Surrogate kernel over $\z$ & Deep kernel RBF \\
\bottomrule
\end{tabular}
\caption{
Main optimization, objective-generation, and modeling settings used in the experiments.}
\label{tab:exp_settings}
\end{table}

\subsection{DGBFGP configuration}\label{sec:dgbfgp_full}
\begin{table}[H]
\begin{center}
\renewcommand{\arraystretch}{1.15}
\begin{tabular}{p{0.32\linewidth}p{0.62\linewidth}}
\hline
\multicolumn{2}{c}{DGBFGP} \\
\hline
Latent dimension &
$d=256$ \\
Max token length &
$L_{\max}=128$ \\
Dataset &
GuacaMol \\
Basis / features &
Hilbert-space features (\texttt{basis\_func=hs}), basis size $M=128$ \\
GP KL scale &
\texttt{gpB}=0.001 (\texttt{gp\_kl\_scale\_B}) \\
Hyperparameters &
batch size $1024$, learning rate $5\times 10^{-4}$ \\
Decoder &
InfoTransformerVAE decoder \\
Dropout &
$p_{\mathrm{drop}}=0.2$ \\
\hline
\end{tabular}
\end{center}
\caption{DGBFGP configuration and training hyperparameters used in the experiments.}
\label{tab:dgbfgpconfig}
\end{table}

\begin{table}[H]
\begin{center}
\begin{tabular}{cl}
\hline
Index & Covariate name \\
\hline
1 & molecular weight \\
2 & number of H-bond donors \\
3 & number of H-bond acceptors \\
4 & number of rotatable bonds \\
5 & Quantitative Estimation of Drug-likeness (QED) \\
6 & Synthetic Accessibility Score (SAS) \\
7 & number of long cycles \\
8 & time covariate $t$ (optional for time version) \\
\hline
\end{tabular}
\end{center}
\caption{Covariates used by \textsc{TALBO} for Multi-property optimization tasks.}
\label{table:cov}
\end{table}

\subsection{InfoTransformerVAE configuration}\label{sec:infovae_config}
\begin{table}[h!]
\begin{center}
\renewcommand{\arraystretch}{1.15}
\begin{tabular}{p{0.32\linewidth}p{0.62\linewidth}}
\hline
\multicolumn{2}{c}{InfoTransformerVAE} \\
\hline
Latent dimension &
$d=256$ (\texttt{bottleneck\_size}=2, \texttt{d\_model}=128) \\
Max decode length &
$L_{\max}=128$  \\
KL weight &
$\beta_{\mathrm{KL}}=0.1$ (\texttt{kl\_factor}) \\
\hline
\end{tabular}
\end{center}
\label{tab:infovae}
\caption{InfoTransformerVAE architecture and training hyperparameters used in the experiments}
\end{table}

\subsection{Computational resources}\label{app:compute}
We run all the experiments on an Nvidia Tesla V100 GPU.

\end{document}